\documentclass[final,5p,times,twocolumn]{elsarticle}
\usepackage[colorlinks=true,linkcolor=blue,anchorcolor=blue,citecolor=blue,filecolor=blue,menucolor=blue,runcolor=blue,urlcolor=blue,breaklinks]{hyperref}
\usepackage[norsk,main=english]{babel}	
\usepackage{soul}  
\usepackage{lipsum} 
\usepackage{placeins}
\usepackage{tabstackengine} 
\usepackage{lettrine} 
\usepackage{amsmath,amsfonts,amssymb} 
\usepackage{amsthm}            
\usepackage[graphicx]{realboxes}
\usepackage{adjustbox}
\usepackage{varwidth}
\usepackage[ruled,vlined]{algorithm2e}
\usepackage{algorithmic}
\newtheorem{definition}{Definition}
\newtheorem{assumption}{Assumption}
\usepackage{subcaption} 
\usepackage{listings}                 
\usepackage{tikzsymbols}              
\usepackage{textcomp}                 
\usepackage{tabulary}
\usepackage{svg}
\usepackage{booktabs}
\usepackage{siunitx}
\usetikzlibrary{decorations.pathreplacing}
\usepackage{tablefootnote} 

\usepackage{afterpage} 

\usepackage[colorinlistoftodos,prependcaption,textsize=tiny]{todonotes}

\usepackage[thinc]{esdiff} 
\usepackage{bm} 
\usepackage{relsize} 
\usepackage{lineno}
\usepackage{float}

\definecolor{green}{rgb}{0,0.6,0}

\DeclareOldFontCommand{\bf}{\normalfont\bfseries}{\mathbf}
\newcolumntype{R}[1]{>{\raggedleft\let\newline\\\arraybackslash\hspace{0pt}}m{#1}}

\usepackage[intoc]{nomencl} 
\usepackage{etoolbox}

\journal{Arxiv}

\newcommand{\bmat}[1]{\begin{bmatrix}#1\end{bmatrix}}

\newcommand{\bsym}[1]{\boldsymbol{#1}}

\newcommand{\pose}{\boldsymbol{\eta}}
\newcommand{\dpose}{\dot{\boldsymbol{\eta}}}
\newcommand{\vel}{\boldsymbol{\nu}}
\newcommand{\dvel}{\dot{\boldsymbol{\nu}}}
\newcommand{\force}{\boldsymbol{\tau}}
\newcommand{\control}{\boldsymbol{u}}
\newcommand{\state}{\boldsymbol{x}}
\newcommand{\pos}{\boldsymbol{p}}
\newcommand{\termat}{\boldsymbol{P}}

\newcommand{\obsv}{\boldsymbol{\zeta}}

\makeatletter
\def\ps@pprintTitle{%
  \let\@oddhead\@empty
  \let\@evenhead\@empty
  \let\@oddfoot\@empty
  \let\@evenfoot\@oddfoot
}
\makeatother

\begin{document}
\begin{frontmatter}
\title{Modular Control Architecture for Safe Marine Navigation: Reinforcement Learning and Predictive Safety Filters}

\author[adiladdress1]{Aksel Vaaler}
\ead{akselva@stud.ntnu.no}

\author[adiladdress1]{Svein Jostein Husa}
\ead{sveinjhu@stud.ntnu.no}

\author[adiladdress1]{Daniel Menges}
\ead{daniel.menges@ntnu.no}

\author[adiladdress1]{Thomas Nakken Larsen}
\ead{thomas.n.larsen@ntnu.no}

\author[adiladdress1,adiladdress2]{Adil Rasheed\corref{mycorrespondingauthor}}
\cortext[mycorrespondingauthor]{Corresponding author}
\ead{adil.rasheed@ntnu.no}

\address[adiladdress1]{Department of Engineering Cybernetics, Norwegian University of Science and Technology}
\address[adiladdress2]{Mathematics and Cybernetics, SINTEF Digital}

\begin{abstract}
Many autonomous systems are safety-critical, making it essential to have a closed-loop control system that satisfies constraints arising from underlying physical limitations and safety aspects in a robust manner. However, this is often challenging to achieve for real-world systems. For example, autonomous ships at sea have nonlinear and uncertain dynamics and are subject to numerous time-varying environmental disturbances such as waves, currents, and wind. There is increasing interest in using machine learning-based approaches to adapt these systems to more complex scenarios, but there are few standard frameworks that guarantee the safety and stability of such systems. Recently, predictive safety filters (PSF) have emerged as a promising method to ensure constraint satisfaction in learning-based control, bypassing the need for explicit constraint handling in the learning algorithms themselves. The safety filter approach leads to a modular separation of the problem, allowing the use of arbitrary control policies in a task-agnostic way. The filter takes in a potentially unsafe control action from the main controller and solves an optimization problem to compute a minimal perturbation of the proposed action that adheres to both physical and safety constraints. In this work, we combine reinforcement learning (RL) with predictive safety filtering in the context of marine navigation and control. The RL agent is trained on path-following and safety adherence across a wide range of randomly generated environments, while the predictive safety filter continuously monitors the agents' proposed control actions and modifies them if necessary. The combined PSF/RL scheme is implemented on a simulated model of Cybership II, a miniature replica of a typical supply ship. Safety performance and learning rate are evaluated and compared with those of a standard, non-PSF, RL agent. It is demonstrated that the predictive safety filter is able to keep the vessel safe, while not prohibiting the learning rate and performance of the RL agent.
\end{abstract}


\end{frontmatter}
\section{Introduction}
\label{sec:Introduction}
The rapid advancement of artificial intelligence (AI) has brought autonomous systems to the forefront of technological development. These systems have transformed various industries, revolutionizing sectors such as transportation, healthcare, and manufacturing. Among the myriad applications, autonomous vessels hold a particularly critical role. Operating in complex and unpredictable marine environments, these vessels face challenges that range from changing weather conditions to potential collisions. Achieving autonomy in such complex systems requires accurate modeling of ship dynamics and its operational environment. However, the development of advanced control algorithms is hindered by incomplete and uncertain models and inputs. In recent years, model-free control algorithms based on reinforcement learning (RL) have gained popularity for their ability to tackle complex tasks. In the context of autonomous vessels, notable studies \cite{Meyer2020ccc, Meyer2020taa, Larsen2021cdr, Heiberg2022rbi} have demonstrated the utility of RL for path-following and collision avoidance tasks. However, training and testing RL algorithms in real environments pose significant safety concerns. As a result, employing RL in systems operating near humans requires critical safety precautions to mitigate the risk of severe damage or loss of life. To address safety concerns, a promising approach involves the application of predictive safety filters (PSF). A PSF acts as a mechanism designed to filter out signals or inputs that could potentially cause harm or damage, thereby ensuring system safety and integrity. The concept of a PSF was first proposed in \cite{Wabersich2021aps} and is based on the model predictive control (MPC) principle \cite{Camacho2007mpc}. It serves as a modular and minimally intrusive safety certification mechanism suitable for various control architectures. Although no previous studies have specifically explored the application of a PSF in learning-based marine craft navigation and control, related methods have been applied to the field of marine vessel collision avoidance (COLAV). For example, Thyri et al. \cite{Thyri2020rca} successfully developed a reactive control module based on the control barrier function (CBF) principle for hazard avoidance in marine vessels. Furthermore, Johansen et al. \cite{Johansen2016sca} proposed a scenario-based model predictive controller (SBMPC) that alters vessel courses to find optimal collision-free paths, demonstrating its effectiveness in collision avoidance.

\textbf{Contribution:} In this work, we propose a hybrid algorithm that combines a predictive safety filter (PSF) and a model-free RL for path-following and collision avoidance tasks, ensuring safe sea operations of autonomous surface vessels (ASVs) in complex environments. The primary contribution of this study lies in the design and verification of a PSF for marine collision avoidance and control. We analyze the performance of the combined PSF/RL scheme by evaluating safety compliance and navigation quality through simulated randomized scenarios with varying difficulty levels. Through a comparison of the performance of RL agents with and without a PSF, we highlight the advantages of incorporating a PSF, both during training and in the test phase. This research aims to make the practical utilization of RL in ASVs more feasible by providing an approach that offers a higher level of transparency and safety assurance compared to existing state-of-the-art RL methods. Moreover, we demonstrate that PSF can speed up learning and reduce training time.

In order to achieve our objectives, this article is structured as follows. Section \ref{sec:theory} presents the essential theory, providing the basis for our approach. The proposed method and the simulation setup are detailed in Section \ref{sec:methodandsetup}. Section \ref{sec:resultsanddiscussions} presents and analyzes the results obtained from our experiments. Finally, Section \ref{sec:conclusionandfuturework} concludes the work and offers insights for future research and considerations.

\section{Theory}
\label{sec:theory}

\subsection{Ship modeling}
To execute the simulations and experiments, Cybership II is used, which is a 1:70 scale replica of a typical supply ship \cite{Skjetne2004mia}. System identification parameters for the vessel were obtained from \cite{Skjetne2004mia}. To simulate the dynamics of the vessel, the three dimensions of freedom (DOF) surge-sway-yaw model described in \cite{Fossen2011hom} was used.

\subsubsection{Kinematics}
We define the state of the system as
\begin{equation}
    \begin{aligned}
        \boldsymbol{\eta} & = \bmat{x & y & \psi}^T, \\
        \boldsymbol{\nu} & = \bmat{u & v & r}^T,
    \end{aligned}
\end{equation}
where $\boldsymbol{\eta}$ denotes the coordinates and heading of the ship respectively, in the north, east, and down (NED) coordinate frame. $\boldsymbol{\nu}$ denotes the angular speed of the vessel's surge, sway and yaw angularity. For a more detailed description of the model kinematics, the the reader is referred to \cite{Fossen2011hom}. The kinematics are defined by \cite{Fossen2011hom}

\begin{equation}\label{eq:kinematics}
    \dpose = \bsym{f}_{kinematic}(\psi,\vel) = \bsym{R}(\psi)\vel,
\end{equation}
where $\bsym{R}(\psi)$ is the rotation matrix expressed by
\begin{equation}
    \bsym{R}(\psi) = \bmat{\cos{\psi} & -\sin{\psi} & 0 \\
                           \sin{\psi} & \cos{\psi} & 0 \\
                           0 & 0 & 1}.
\end{equation}

\subsubsection{dynamics}
The model dynamics are defined according to \cite{Skjetne2004mia}
\begin{equation}
    \bsym{M}\dvel + \bsym{C}(\vel)\vel + \bsym{D}(\vel)\vel = \force + \force_d,
\end{equation}
where $\bsym{M}$ denotes the mass matrix
\begin{equation}
    \bsym{M} = \bmat{m_{11} & 0 & 0 \\
                     0 & m_{22} & m_{23} \\
                     0 & m_{32} & m_{33}},
\end{equation}
$ \bsym{C}(\vel)$ is the Coriolis matrix
\begin{equation}
    \begin{aligned}
        c_{13} & = -m_{11}v - m_{23}r, \\
        c_{23} & = m_{11}u, \\
        \bsym{C}(\vel) & = \bmat{0 & 0 & c_{13} \\
                                 0 & 0 & c_{23} \\
                                 -c_{13} & -c_{23} & 0},
    \end{aligned}
\end{equation}
and $\bsym{D}(\vel)$ characterizes the damping matrix
\begin{equation}
    \begin{aligned}
        d_{11} & = -X_u - X_{|u|u}|u| - X_{uuu}u^2, \\
        d_{22} & = -Y_v - Y_{|v|v}|v| - Y_{|r|v}|r|, \\
        d_{23} & = -Y_r - Y_{|v|r}|v| - Y_{|r|r}|r|, \\
        d_{32} & = -N_v - N_{|v|v}|v| - N_{|r|v}|r|, \\
        d_{33} & = -N_r - N_{|v|r}|v| - N_{|r|r}|r|, \\
        \bsym{D}(\vel) & = \bmat{d_{11} & 0 & 0 \\
                                 0 & d_{22} & d_{23} \\
                                 0 & d_{32} & d_{33}}.
    \end{aligned}
\end{equation}
Furthermore, $\force = \bmat{F_u, F_v, T_r}^T$ denotes the control input containing the surge force, sway force, and yaw moment applied to the ship.
Since the mass matrix $\bsym{M}$ is invertible, we can rewrite the dynamic equations as the explicit ODE
\begin{equation}\label{eq:dynamics}
    \dvel = \bsym{f}_{dynamic}(\vel,\force) = \bsym{M}^{-1}(-\bsym{C}(\vel)\vel - \bsym{D}(\vel)\vel + \force + \force_d),
\end{equation}
where $\force_d$ is a vector of generalized disturbance forces affecting the ship.

\subsection{Collision risk}
The collision risk index (CRI) is a metric used to quantify the risk of collision between two vessels. The approach used in this work is based on \cite{Heiberg2022rbi}, which is designed to be compatible with the simulation environment. The CRI is calculated by the estimated time to the closest point of approach (TCPA) and the estimated distance to the closest point of approach (DCPA). Additionally, it considers the relative distance $R$, the relative speed $V$, and the relative bearing $\theta_T$. For this purpose, it uses fuzzy comprehensive evaluation methods, where the membership functions $u(\cdot) \in[0,1]$ are used to determine the risk level associated with each risk factor. Consequently, the CRI is given by
\begin{align}
\mathrm{CRI}= &\ \alpha_{\mathrm{CPA}} \sqrt{u(\mathrm{DCPA}) \cdot u(\mathrm{TCPA})} \\
 &+\alpha_{\theta_{\mathrm{T}}} u\left(\theta_{\mathrm{T}}\right)+\alpha_{\mathrm{R}} u(R)+\alpha_{\mathrm{V}} u(V), \nonumber
\end{align}
where $\alpha_{\mathrm{DCPA}}+\alpha_{\mathrm{TCPA}}+\alpha_{\theta_{\mathrm{T}}}+\alpha_{\mathrm{R}}+\alpha_{\mathrm{V}}=1$. Further details on the CRI calculation can be found in \cite{Heiberg2022rbi}. 

\subsection{Deep reinforcement learning}
In this section, we introduce the field of deep reinforcement learning (DRL) to provide an understanding of the RL methods used in this work. RL is a subfield of machine learning (ML) where a sequential decision-making agent learns a behavioral policy by iteratively acting in its environment and optimizing its parameters for the reward it receives. The term ``deep'' refers to parameterizing the agent using deep neural networks (DNN), improving the applicability and performance of RL methods, particularly in complex environments with high-dimensional state spaces \cite{Degrave2022mcoa, Silver2017mca}.

\subsubsection{RL preliminaries}
One core assumption for RL is that the environment can be modeled as a Markov Decision Process (MDP), which is defined by the tuple $\left\{\mathcal{S}, \mathcal{A}, \mathcal{T}, \mathcal{R}, p\left(s_0\right), \gamma\right\}$, where
\begin{itemize}
    \item $\mathcal{S}$ is the state space,
    \item $\mathcal{A}$ is the action space,
    \item $\mathcal{T}: \mathcal{S} \times \mathcal{A} \rightarrow p(\mathcal{S})$ is the transition function that represents the probability that an agent ends up in a new state $s^{\prime} \in \mathcal{S}$ after taking a specific action $a \in \mathcal{A}$ from a state $s \in \mathcal{S}$,
    \item $r: \mathcal{S} \times \mathcal{A} \times \mathcal{S} \rightarrow \mathbb{R}$ is the reward function, which returns a scalar reward associated with the transition function,
    \item $p\left(s_0\right)$ is the initial state distribution, and
    \item $\gamma \in[0,1]$ is the discount factor. 
\end{itemize}
An agent acts in an environment according to a policy $\pi:~\mathcal{S}~\rightarrow~p(\mathcal{A})$. Solving the MDP is equivalent to finding an optimal policy $\pi^{\star}$ that maximizes the reward. Therefore, we need to introduce the action-value function $Q^\pi(s, a)$, defined as the expectation of the cumulative return for a given policy $\pi$ dependent on the state $s$ and the action $a$. As a result, the action-value function at time $t$ under consideration of a specific window size $K$ is expressed by
\begin{equation}
    Q^\pi(s, a)=\mathbb{E}_{\pi, \mathcal{T}}\left[\sum_{k=0}^K \gamma^k r_{t+k} \mid s_t=s, a_t=a\right].
\end{equation}
The optimal policy $\pi^{\star}$ can be defined as the policy that maximizes the expected return of $Q^\pi(s, a)$.
\begin{align}
    \pi^*&=\underset{\pi}{\arg \max } ~Q^\pi(s, a)\\
    &=\underset{\pi}{\arg \max } ~\mathbb{E}_{\pi, \mathcal{T}}\left[\sum_{k=0}^K \gamma^k r_{t+k} \mid s_t=s, a_t=a\right]
\end{align}

\subsubsection{Policy-based and value-based Methods}
There are two main approaches to train model-free RL agents; \textit{value-based methods} learn an action-value function $Q^\pi(s, a)$ and derive the optimal policy from it, while \textit{policy-based methods} directly optimize a policy function $\pi_\theta(a \mid s)$ with parameters $\theta$. Modern RL algorithms assume a hybrid approach, \textit{actor-critic methods}, fitting both a value function and an explicit policy to balance a bias-variance trade-off. While the value-based approach is often more sample-efficient, it also tends to be less stable \cite{Nachum2017btg}. Examples of value-based methods algorithms are Deep-Q Networks (DQN) \cite{Mnih2013paw} and Hindsight Experience Replay (HER) \cite{Andrychowicz2017her}. Policy-based methods directly optimize an explicitly parameterized policy and are generally more stable than value-based methods. However, this stability comes at the cost of high sample complexity due to the optimization's on-policy property. REINFORCE algorithms \cite{williams1992simple} are examples of pure policy gradient methods. Modern policy gradient methods often focus on optimizing the Advantage function, which entails fitting a state-value function (critic) to approximate the advantage for use in the policy (actor) gradient loss. Such \textit{actor-critic} algorithms overcome many of the limitations of policy-based and value-based methods \cite{Konda1999aca}. Examples of actor-critic methods are Deep Deterministic Policy Gradient (DDPG) \cite{Lillicrap2019ccw}, Soft Actor-Critic (SAC) \cite{Haarnoja2018sac}, Trust Region Policy Optimization (TRPO) \cite{Schulman2015trp} and Proximal Policy Optimization (PPO) \cite{Schulman2017ppo}.

\subsubsection{Proximal policy optimization}
Proximal policy optimization (PPO) is a popular model-free algorithm in the family of policy gradient methods, originally developed by Shulman et al. \cite{Schulman2017ppo}. Employing a trust region ensures that new policies do not deviate far from the old policy, avoiding large policy updates that could result in unstable learning. It does this by using a clipped surrogate objective function, which is a modified version of the standard policy gradient objective. The surrogate objective can be defined as 

\begin{equation}
L^{C P I}(\theta)=\hat{\mathbb{E}}_t\left[\frac{\pi_\theta\left(a_t \mid s_t\right)}{\pi_{\theta_{\text {old }}}\left(a_t \mid s_t\right)} \hat{A}_t\right]=\hat{\mathbb{E}}_t\left[r_t(\theta) \hat{A}_t\right],
\end{equation}
where $\hat{A}_t$ is the advantage estimate, which is a measure of how much better an action is compared to the average action in a given state, and $\theta$ is a set of parameters defining a policy $\pi$. To avoid large policy updates, the PPO algorithm has a clipped surrogate objective function, which is expressed by

\begin{equation}
    L^{C L I P}(\theta)=\hat{\mathbb{E}}_t\left[\min \left(r_t(\theta) \hat{A}_t, \operatorname{clip}\left(r_t(\theta), 1-\epsilon, 1+\epsilon\right) \hat{A}_t\right)\right],
\end{equation}
ensuring that the probability ratio $r_t(\theta)$ is within the range $1-\epsilon$ and $1+\epsilon$.

The PPO algorithm is considered to be an efficient and robust algorithm, suited for various domains \cite{Schulman2017ppo}, and has shown good results in previous work on the \textbf{gym-auv} simulation environment \cite{Meyer2020ccc, Heiberg2022rbi}. Most notably, in \cite{Larsen2021cdr} the PPO algorithm was shown to outperform other state-of-the-art algorithms in a range of different scenarios in the \textbf{gym-auv} environment. Therefore, PPO was chosen as the RL method for this work. 

\subsection{Environmental disturbance observer}
To estimate the environmental disturbances affecting the ship, we use the nonlinear disturbance observer algorithm developed in \cite{Menges2023aed}. The observer system is defined by
\begin{equation}\label{eq:observer_system_general}
    \begin{aligned}
        \hat{\force}_d & = \obsv + \bsym{\mu}(\vel), \\
        \dot{\obsv} & = \bsym{h}(\vel,\hat{\force}_d),
    \end{aligned}
\end{equation}
where $\hat{\force}_d$ is the estimate of the environmental disturbance forces affecting the ship, and $\obsv$ is an observer variable. The estimated error and error dynamics are then described by
\begin{equation}
    \begin{aligned}
        \bsym{z} & = \force_d - \obsv - \bsym{\mu}(\vel), \\
        \dot{\bsym{z}} & = \dot{\force}_d - \dot{\obsv} - \frac{\partial \bsym{\mu}}{\partial \vel}\dot{\vel}.
    \end{aligned}
\end{equation}
Assuming that the true disturbance $\force_d$ is relatively slow-varying $(\dot{\force_d} \approx \bsym{0})$ and inserting the ship model dynamics, the error dynamics become
\begin{equation}
    \dot{\bsym{z}} = -\bsym{h}(\vel,\hat{\force}_d) - \frac{\partial \bsym{\mu}}{\partial \vel} \bsym{M}^{-1}(\bsym{D}(\vel)\vel - \bsym{C}(\vel)\vel + \force + \force_d).
\end{equation}
By defining $\bsym{h}(\vel,\hat{\force}_d)$ as
\begin{equation}\label{eq:observer_h_def}
    \bsym{h}(\vel,\hat{\force}_d) = - \frac{\partial \bsym{\mu}}{\partial \vel} \bsym{M}^{-1}(\bsym{D}(\vel)\vel - \bsym{C}(\vel)\vel + \force + \hat{\force}_d),
\end{equation}
authors in \cite{Menges2023aed} show that the observer error converges by applying
\begin{equation}\label{eq:observer_T}
    \frac{\partial \bsym{\mu}}{\partial \vel} = \bsym{T} = \bmat{\Gamma_1 \frac{1}{k_{11}} \sigma & 0 & 0 \\[6pt]
                                                                 0 & \Gamma_2 \frac{1}{k_{22}} & -\Gamma_2 \frac{k_{23}}{k_{22}k_{33}} \\[6pt]
                                                                 0 & -\Gamma_3 \frac{k_{32}}{k_{22}k_{33}} & \Gamma_3 \frac{1}{k_{33}}},
\end{equation}
where $\sigma = 1 - \frac{k_{23}k_{32}}{k_{22}k_{33}}$, $k_{ij}$ are the elements of the inverse mass matrix
\begin{equation}
    \bsym{M}^{-1} = \bmat{k_{11} & 0 & 0 \\
                          0 & k_{22} & k_{23} \\
                          0 & k_{32} & k_{33}},
\end{equation}
and $\Gamma_{1,2,3}$ are adaptive gains, which will be chosen later. Using equations \eqref{eq:observer_h_def} and \eqref{eq:observer_T}, the disturbance observer system \eqref{eq:observer_system_general} becomes
\begin{equation}\label{eq:observer_system_final}
    \begin{aligned}
        \hat{\force}_d & = \obsv + \bsym{T}\vel, \\
        \dot{\obsv} & = -\bsym{T} \bsym{M}^{-1}(\bsym{D}(\vel)\vel - \bsym{C}(\vel)\vel + \force + \hat{\force}_d). 
    \end{aligned}
\end{equation}

\subsection{Predictive Safety Filter}
To ensure safe exploration, a predictive safety filter (PSF) is integrated into the control loop of the system. The main idea of the PSF, as first described in \cite{Wabersich2021aps},  is to predict the future states of the system based on the current system state $\state(t)$ and the current proposed control action of the RL agent $\control_L(t)$, and find the minimally perturbed control action $\control_0^*(\state(t), \control_L(t))$, which guarantees a safe state trajectory for all times $t' \in (t,\infty)$.
At every time step the PSF solves the finite-horizon optimal control problem (OCP): 
\begin{equation}
    \begin{aligned}
        \min_{\control_{i|k}, \state_{i|k}} & ||\control_{0|k} - \control_L(k)||_W^2 \\
        s.t. \quad & (a) \quad \state_{0|k} = \state(k) \\
        & (b) \quad \state_{i+1|k} = \bsym{f}(\state_{i|k}, \control_{i|k}, \Delta T) \quad \forall i \in [0,N - 1]\\
        & (c) \quad \state_{i|k} \in \mathbb{X} \quad \forall i \in [0,N] \\
        & (d) \quad \control_{i|k} \in \mathbb{U} \quad \forall i \in [0,N - 1] \\
        & (e) \quad \state_{N|k} \in \mathbb{X}_f
    \end{aligned}
    \label{eq:general_PSF}
\end{equation}

Here, $\state(k)$ and $\control_L(k)$ are the system states and the proposed RL control action at the current time step, respectively. $W$ is the weighting matrix for the cost function and $\bsym{f}(\cdot)$ are the discretized system model equations, where $\Delta T$ is the discretization step size. $N$ is the number of states in the predicted trajectory (shooting nodes), which means that the prediction horizon is given by $T_f = N\Delta T$. 

$\mathbb{X} \subseteq \mathbb{R}^{n_x}$ denotes the set of feasible (safe) states, while $\mathbb{U} \subseteq \mathbb{R}^{n_u}$ is the set of feasible control inputs, where $n_x$ and $n_u$ are the dimensions of the states and control inputs, respectively. The set $\mathbb{X}_f \subseteq \mathbb{X}$ is called the \textit{terminal safe set} \cite{Wabersich2021aps}. The terminal safe set is a \textit{control invariant set}, which is deinfed as follows:
\begin{definition}
(Control invariant set). The set $\mathbb{X}_f$ is a control invariant set if and only if, for all $\state|(\state \in \mathbb{X}_f)$, there exists a control law $\control = \bsym{k}(\state) \subseteq \mathbb{U}$ such that $\bsym{f}(\state,\bsym{k}(\state)) \in \mathbb{X}_f$.
\end{definition}
The terminal safety constraint $\state_{N|k} \in \mathbb{X}_f$ thus guarantees that the system will be able to stay within the terminal safe set $\mathbb{X}_f$ (and, by extension, the feasible set $\mathbb{X}$) for all time $t' \in (T_f, \infty)$.

\begin{figure}[htb]
    \begin{subfigure}{0.495\linewidth}
        \includegraphics[width=\linewidth]{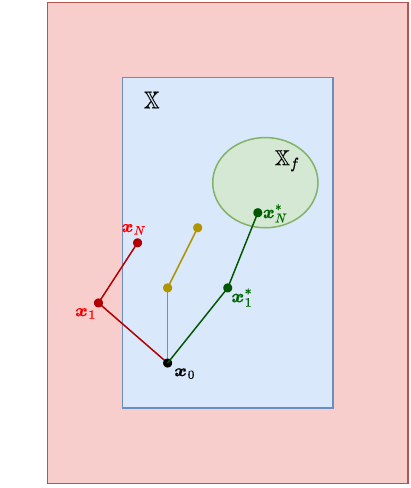}
        \caption{PSF trajectory modification with N = 2}
    \end{subfigure}
    \begin{subfigure}{0.495\linewidth}
        \includegraphics[width=\linewidth]{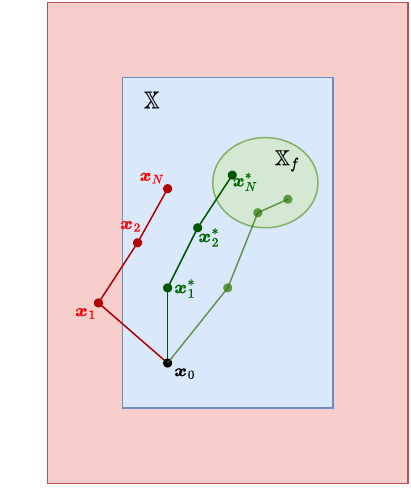}
        \caption{PSF trajectory modification with N = 3}
    \end{subfigure}
    \caption[Illustration of predictive safety filter mechanism]{Illustration of how the predictive safety filter modifies the nominal trajectory based on a safe set $\mathbb{X}$, terminal set $\mathbb{X}_f$, and number of shooting nodes $N$. The yellow path indicated on the left figure lies closer to the nominal unsafe path (red), but given the short prediction horizon, the PSF must take the shorter (dark green) path directly towards the terminal set. In the righthand figure, with 1 additional shooting node, the PSF can compute a trajectory that lies closer to the nominal path, and still be able to reach the terminal set ``in time"}
    \label{fig:PSF_illustration}
\end{figure}

After solving the OCP, the control action that is applied to the system is chosen as $\control_0^*(\state(k),\control_L(k)) = \control_{0|k}^*$. The PSF algorithm is similar to the classic MPC algorithm \cite{Camacho2007mpc}. The difference is that while classical MPC minimizes the OCP with respect to a reference \textit{trajectory}, the PSF only minimizes with respect to a reference \textit{control input}. Fig.~\ref{fig:PSF_illustration} gives a graphical explanation of the working of PSF.

\subsubsection{Formulation of PSF OCP for the 3-DOF vessel model}
Recall the definition of the PSF OCP:
\begin{equation}
    \begin{aligned}
        \min_{\control_{i|k}, \state_{i|k}} & ||\control_{0|k} - \control_L(k)||_W^2 \\
        s.t. \quad & (a) \quad \state_{0|k} = \state(k) \\
        & (b) \quad \state_{i+1|k} = \bsym{f}(\state_{i|k}, \control_{i|k}, \Delta T) \quad \forall i \in [0,N - 1]\\
        & (c) \quad \state_{i|k} \in \mathbb{X} \quad \forall i \in [0,N] \\
        & (d) \quad \control_{i|k} \in \mathbb{U} \quad \forall i \in [0,N - 1] \\
        & (e) \quad \state_{N|k} \in \mathbb{X}_f
    \end{aligned}
    \tag{\eqref{eq:general_PSF} revisited}
\end{equation}
Replacing the state variable $\state$ with the vessel's pose $\pose$ and velocities $\vel$, constraints \textit{(a)} and \textit{(b)} can be reformulated as
\begin{equation}\label{eq:model_constraints}
    \begin{aligned}
        & \pose_{0|k} = \pose(k), \\
        & \vel_{0|k} = \vel(k), \\
        & \pose_{i+1|k} = \bsym{f}_{kinematic}(\psi_{i|k}, \vel_{i|k}, \Delta T) \quad \forall i \in [0,N - 1],\\
        & \vel_{i+1|k} = \bsym{f}_{dynamic}(\vel_{i|k}, \control_{i|k}, \Delta T) \quad \forall i \in [0,N - 1].\\
    \end{aligned}
\end{equation}
The state constraint $\state_{i|k} \in \mathbb{X}$ is deemed equivalent to the following:
\begin{enumerate}
    \item All ship velocities are within specified upper and lower bounds: $\vel_{lb} \leq \vel_{i|k} \leq \vel_{ub}$
    \item The position of the vessel is a safe distance away from any observed obstacles: $d(\pos_{i|k},\mathbb{O}_{i|k}) \geq d_{safe}$
\end{enumerate}
Note that $d(\cdot)$ is the Euclidean distance function, $\pos_{i|k} = \bmat{x & y}^T$ is the NED coordinate position of the vessel at prediction step $i$ and time step $k$, $\mathbb{O}_{i|k} \subseteq \mathbb{R}^2$ is the union of all observed obstacles, and $d_{safe}$ is the minimum safety distance. This can be stated formally as
\begin{equation} \label{eq:ship_state_constraints}
    \begin{aligned}
        \state_{i|k} & \in \mathbb{X} \rightarrow \pose_{i|k} \in \mathbb{X}_{\eta} \cap \vel_{i|k} \in \mathbb{X}_{\nu}, \\
        \pose_{i|k} & \in \mathbb{X}_{\eta} \leftrightarrow d(\pos_{i|k},\mathbb{O}_{i|k}) \geq d_{safe}, \\
        \vel_{i|k} & \in \mathbb{X}_{\nu} \leftrightarrow \vel_{lb} \leq \vel_{i|k} \leq \vel_{ub},
    \end{aligned}
\end{equation}
and the set $\mathbb{U}$ is defined by
\begin{equation}
    \control_{i|k} \in \mathbb{U} \leftrightarrow \control_{lb} \leq \control_{i|k} \leq \control_{ub}.
\end{equation}
The implementation of the collision avoidance constraint will be described in Section \ref{sec:methodandsetup}.

\subsubsection{Control invariant terminal set formulation}
The formulation of the terminal safe set $\mathbb{X}_f$ generally follows the procedure used in \cite{Tearle2021aps}. The main difference is that while \cite{Tearle2021aps} formulates the terminal safe set with respect to lateral track error and relative track heading error, we have chosen to formulate $\mathbb{X}_f$ only with respect to velocities $\vel$, which ensures asymptotic safety while retaining maximal flexibility with respect to the proposed actions of the RL agent.

\begin{assumption} \label{ass:control_inv_ell_set}
    (Control invariant ellipsoidal set) Let the ellipsoidal set $\mathbb{X}_e \subseteq \mathbb{X}$ be defined by $\mathbb{X}_e := \{ \state | \state^T\termat\state \leq 1 \}$ where $\termat$ is a positive definite matrix. Assume that there exists a control law $\bsym{k}(\state) \in \mathbb{U} | (\state \in \mathbb{X}_e)$ and $\termat$ such that $\state_{i|k} \in \mathbb{X}_e \rightarrow \bsym{f}(\state_{i|k},\bsym{k}(\state_{i|k})) \in \mathbb{X}_e$.
\end{assumption}

Assumption \ref{ass:control_inv_ell_set} implies the existence of a control law $\bsym{k}(\state)$ and positive definite matrix $\termat$ which guarantees that the state $\state$ will remain in $\mathbb{X}_e \subseteq \mathbb{X}$ for $t \in (t',\infty)$ as long as $\state(t') \in \mathbb{X}_e$. We first define the terminal collision avoidance constraint
\begin{equation} \label{eq:term_col_avoid}
    d(\pos_{N|k},\mathbb{O}_{N|k}) \geq d_{safe} + d_{f},
\end{equation}
where $(\cdot)_{N|k}$ denotes terminal state, and $d_f$ is an additional safety buffer. Let us now assume that there exists a positive definite matrix $\bsym{P}_f$ and control law $\bsym{k}_f(\state)$ which satisfy assumption \ref{ass:control_inv_ell_set} for the set
\begin{equation}
    \mathbb{C}_f := \{\state| \; ||\pos|| \leq d_f \; \cap \; \vel_{lb} \leq \vel \leq \vel_{ub} \}
    \label{eq:terminal-feasible-set}
\end{equation}
where $||\pos||$ is the euclidean distance from the coordinates $\pos$ to the origin. This means that
\begin{equation} \label{eq:Pf_equation}
    \begin{aligned}
        & \state_{j|k}^T\bsym{P}_f \state_{j|k} \leq 1 \\
        & \rightarrow \bsym{f}(\state_{j|k}, \bsym{k}_f(\state_{j|k}))^T\bsym{P}_f \bsym{f}(\state_{j|k}, \bsym{k}_f(\state_{j|k})) = \state_{j+1|k}^T\bsym{P}_f \state_{j+1|k} \leq 1 \\
        & \rightarrow \state_{j+1|k} \in \mathbb{C}_f
    \end{aligned}
\end{equation}
Defining the variable transformation $\bar{\state}_{j|k} = [\pose_{j|k} - \pose_{N|k},\: \vel_{j|k}]^T$ and applying \ref{eq:Pf_equation} recursively yields:
\begin{equation} \label{eq:Pf_transformed}
    \begin{aligned}
        & \bar{\state}_{j|k}^T \bsym{P}_f \bar{\state}_{j|k} \leq 1 \\
        & \rightarrow \bar{\state}_{\infty|k}^T \bsym{P}_f \bar{\state}_{\infty|k} \leq 1 \\
        & \rightarrow ||\pos_{\infty|k} - \pos_{N|k}|| \leq d_f
    \end{aligned}    
\end{equation}
Inserting $\bar{\state}_{N|k} = [\bsym{0},\: \vel_{N|k}]$ and combining \ref{eq:term_col_avoid} with \ref{eq:Pf_transformed} gives the full terminal constraint formulation

\begin{equation}\label{eq:terminal_safety_constraint}
    \begin{aligned}
        & \bmat{-d(\pos_{N|k},\mathbb{O}_{N|k}) \\
        \bar{\state}_{N|k}^T\bsym{P}_f\bar{\state}_{N|k}} \leq \bmat{-(d_{safe} + d_f) \\ 1}
        \\ & \rightarrow  \quad
        \bmat{-d(\pos_{N|k},\mathbb{O}_{N|k}) \\
        \bmat{0 & \vel_{N|k}}\bsym{P}_f \bmat{0 \\ \vel_{N|k}}} \leq \bmat{-(d_{safe} + d_f) \\ 1}
        \\ & \rightarrow \quad
        \bmat{-d(\pos_{N|k},\mathbb{O}_{N|k}) \\
        \vel_{N|k}^T\bsym{P}_{f_{\vel}} \vel_{N|k}} \leq \bmat{-(d_{safe} + d_f) \\ 1},
    \end{aligned}
\end{equation}
where
\begin{equation}
    \bsym{P}_{f_{\vel}} = \bmat{p_{44} & p_{45} & p_{46} \\
                                p_{54} & p_{55} & p_{56} \\
                                p_{64} & p_{65} & p_{66}}.
\end{equation}

\begin{figure}
    \centering
    \includegraphics[width=0.5\linewidth]{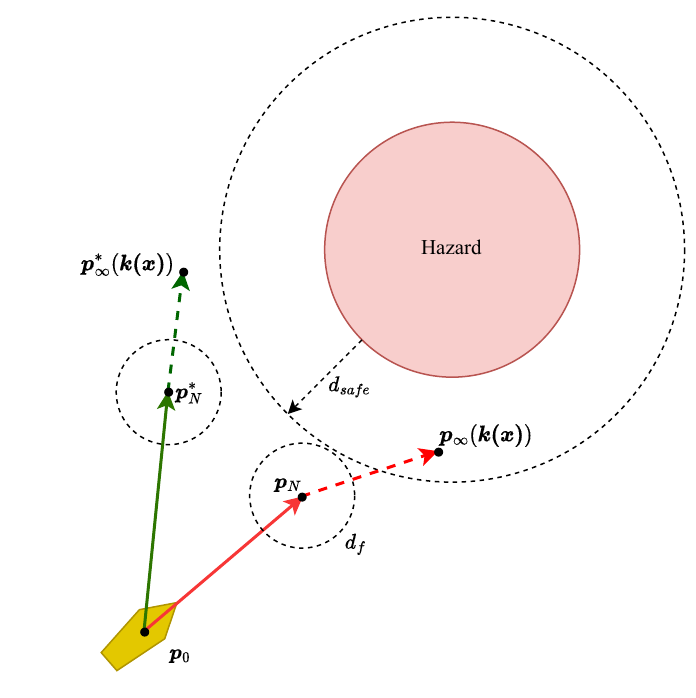}
    \caption[Terminal safety constraint visualization]{Visualization of ship trajectory modification caused by terminal safety constraint, with $N=1$ for clarity. Red arrows indicate nominal (unsafe) trajectory, while green arrows indicate PSF modified trajectory}
    \label{fig:ship_PSF_visual}
\end{figure}

Equation \eqref{eq:terminal_safety_constraint} ensures, given the terminal velocities $\vel_{N|k}$ and the control law $\bsym{k}_f(\state_f)$, that the position of the ship beyond the prediction horizon ($\pos_{\infty}$) will deviate no more than a distance $d_f$ from the terminal position $\pos_{N|k}$. From this, we can conclude that:
\begin{equation}
    \begin{aligned}
         ||\pos_{\infty} - \pos_{N|k}||_2 \leq d_{f} \quad \cap & \quad d(\pos_{N|k},\mathbb{O}_{N|k}) \geq d_{safe} + d_{f}, \\
         & \quad d(\pos_{\infty},\mathbb{O}_{N|k}) \geq d_{safe}.
    \end{aligned}
\end{equation}
The computation of the terminal control law $\bsym{k}_f(\state_f)$ and matrix $\bsym{P}_f$ is described in Section \ref{sec:methodandsetup}. Fig.~\ref{fig:ship_PSF_visual} illustrates a simple scenario in which the PSF modifies the trajectory of the ship due to the terminal safety constraint. The red arrows indicate the nominal path of the ship. While the nominal predicted position $\pos_1$ satisfies the distance requirement $d_{safe}$, the point $\pos_{\infty}(\bsym{k(x)})$ indicates that given the pose and velocity at $\pos_1$, it is not possible to avoid a future safety violation given the terminal control law $\bsym{k(x)}$ (which implies that the constraint of the terminal set is violated). Therefore, the PSF modifies the control input so that the optimal (predicted) safe path is taken instead, indicated by the green arrows. 

Fig.~\ref{fig:ship_PSF_visual} gives an illustration of the modification caused by terminal safety constraint, with $N=1$. The full PSF optimal control problem formulation for the 3-DOF ship model thus becomes:
\begin{equation}
    \begin{aligned}
        \min_{\control_{i|k}, \pose_{i|k}, \vel_{i|k}} & ||\control_{0|k} - \control_L(k)||_W^2 \\
        s.t. \quad & (a) \quad \pose_{0|k} = \pose(k) \\
        & (b) \quad \vel_{0|k} = \vel(k) \\
        & (c) \quad \pose_{i+1|k} = \bsym{f}_{kinematic}(\psi_{i|k}, \vel_{i|k}, \Delta T) \quad \forall i \in [0,N - 1]\\
        & (d) \quad \vel_{i+1|k} = \bsym{f}_{dynamic}(\vel_{i|k}, \control_{i|k} \Delta T) \quad \forall i \in [0,N - 1]\\
        & (e) \quad d(\pos_{i|k},\mathbb{O}_{i|k}) \geq d_{safe} \quad \forall i \in [0,N] \\
        & (f) \quad d(\pos_{N|k},\mathbb{O}_{N|k}) \geq d_{safe} + d_{f} \\
        & (g) \quad \vel_{lb} \leq \vel_{i|k} \leq \vel_{ub} \quad \forall i \in [0,N] \\
        & (h) \quad \control_{lb} \leq \control_{i|k} \leq \control_{ub} \quad \forall i \in [0,N - 1] \\
        & (i) \quad \vel_{N|k}^T\bsym{P}_{f_{\vel}}\vel_{N|k} \leq 1
    \end{aligned}
    \label{eq:ship_PSF}
\end{equation}

\section{Method and setup}
\label{sec:methodandsetup}

\subsection{RL/PSF overview}
\begin{figure}[htb]
    \centering
    \includegraphics[width=\linewidth]{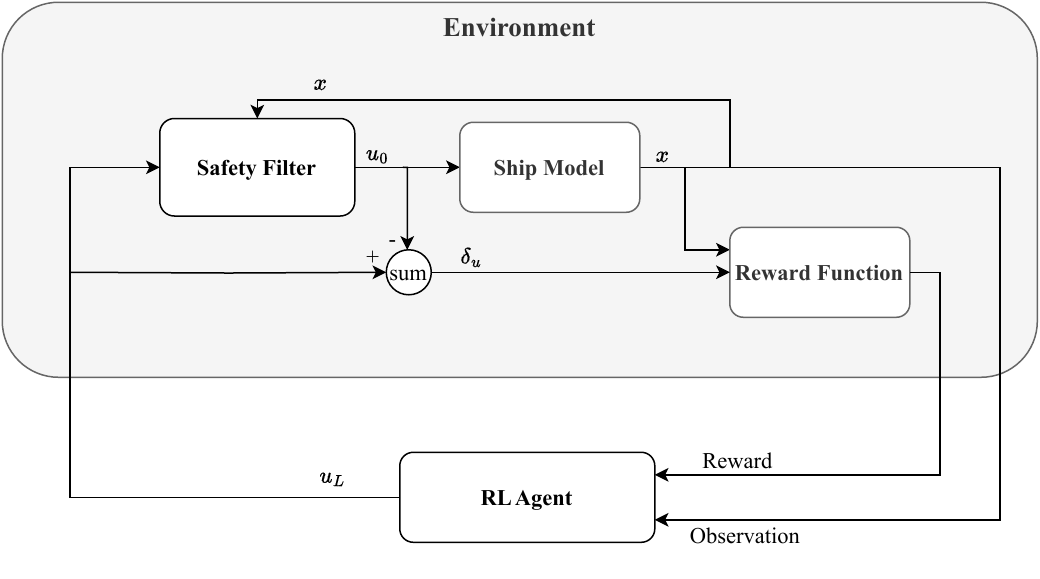}
    \caption{Illustration of the RL + PSF control design. Note that the LiDAR perception features, environmental disturbances, and obstacles have been omitted from this figure for clarity.}
    \label{fig:RL_PSF_fig}
\end{figure}

An overview of the RL/PSF control architecture is shown in Fig.~\ref{fig:RL_PSF_fig}. Based on the current reward and observation (navigation and perception features), the RL agent proposes a control action. The proposed action $u_L$, along with the system state is passed to the PSF, which computes the minimally modified safe control action $u_0$. Subsequently, $u_0$ is passed as the control input for the next iteration of the model simulation. The difference between the proposed action $u_L$ and the modified action $u_0$ is that $\delta_u$ is propagated to the reward function, along with the observation vector.

\subsection{Training environment}
The \textbf{gym-auv} simulation framework by Meyer \cite{Meyer2020ccc, Meyer2020taa} was used to train and test the RL agents. The framework is based on OpenAI Gym \cite{Brockman2016og}, a widely used toolkit to develop and compare RL algorithms. The RL agent is trained using the Stable-Baselines3 PPO implementation \cite{Raffin2021sbr} with hyperparameters identical to those used by Larsen et al. \cite{Larsen2021cdr}. Fig.~\ref{fig:RL_PSF_fig} shows a schematic overview of the control design. 

\subsection{Observation vector}
\begin{figure}
    \centering
    \includegraphics[width=\linewidth]{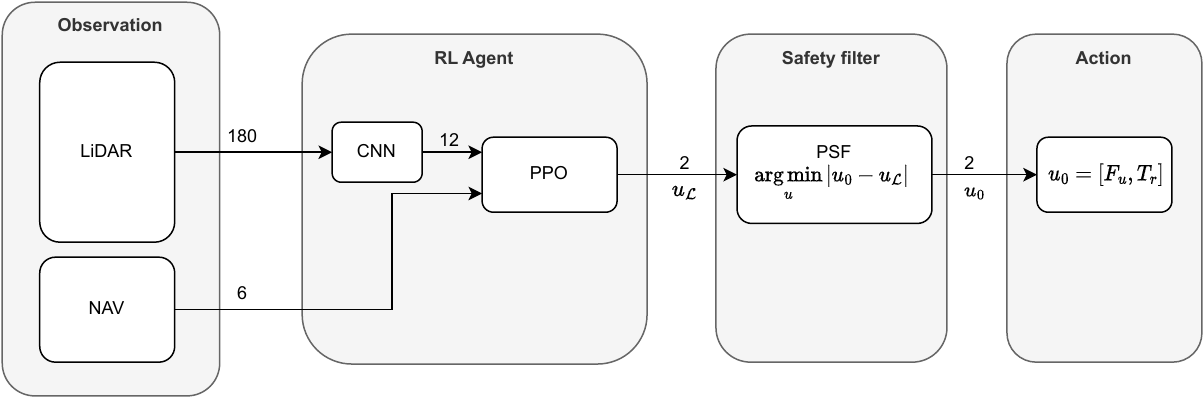}
    \caption[RL agent diagram]{RL agent diagram. The observation vector contains both LiDAR and navigation features. While the navigation features are used directly in the PPO algorithm, the LiDAR measurements are processed through a CNN. The PPO outputs an action $u_{\mathcal{L}}$ that is sent through the safety filter. Finally, the safe action $u_{0}$ can be executed in the environment.}
    \label{fig:RL_agent_diagram}
\end{figure}

The simulated vessel provides both navigation features regarding the path and LiDAR-based distance measurements for the training of the RL agent. Furthermore, based on the paper by Larsen et al. \cite{Larsen2022CNNrisk}, the LiDAR measurements are encoded by a pre-trained convolutional neural network (CNN) to predict the CRI corresponding to the measurements. Further details on the architecture and training of the CNN can be found in \cite{Larsen2022CNNrisk}. Fig.~\ref{fig:RL_agent_diagram} outlines how the observation vector is processed through the RL agent and the PSF.

\subsubsection{Integrated LiDAR sensor suite}
\begin{figure}[h!]
\centering
\begin{subfigure}[h]{0.49\linewidth}
\centering
\includegraphics[width=\linewidth, height=0.7\linewidth]{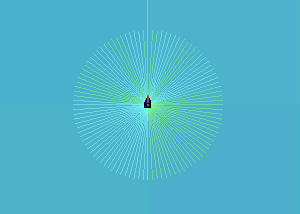}
\subcaption{Visualization of LiDAR detection rays with $N_{rays} = 100$ and $R_{detect} = 50$}
\label{fig:lidar_sensor}
\end{subfigure}
\begin{subfigure}[h]{0.49\linewidth}
\centering
\includegraphics[width=\linewidth, height=0.7\linewidth]{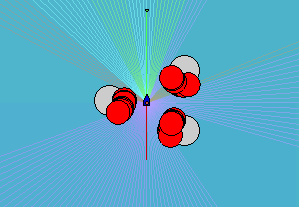}
\subcaption{Visualization of obstacles constructed for PSF using lidar detection}
\label{fig:PSF_lidar_obstacles}
\end{subfigure}
\caption{Schematic of how the LiDAR detection works}
\label{fig:LIDAR}
\end{figure}

\textbf{gym-auv} features an integrated 2D LiDAR sensor suite which is used to detect potential hazards in the vicinity of the ownship. The simulated 2D LiDAR sensor consists of $N_{ray}$ evenly spaced detection rays, each measuring the closest distance to an object along the direction of the ray, within the maximum detection distance $R_{detect}$. The rays are divided into $N_{sector}$ non-overlapping sectors, each sector containing $\frac{N_{ray}}{N_{sector}}$ detection rays. A visualization of the LiDAR detection rays and obstacles is shown in Fig.~\ref{fig:LIDAR}, and the used LiDAR parameters are presented in Table~\ref{tab:lidar-params}.

\begin{table}[h]
\centering
\caption{LiDAR sensor parameters}
\label{tab:lidar-params}
\begin{tabular}{c c c}
\hline
\textbf{Parameter} & \textbf{Description} & \textbf{Value} \\
\hline
$N_{ray}$ & Number of detection rays & 180 \\
$N_{sector}$ & Number of detection sectors  &  20\\
$R_{detect}$ & Maximum detection distance  & 150 ($m$) \\

\end{tabular}
\end{table}


\subsection{Ship model parameters}
The model parameters used for the 3-DOF Cybership II model were obtained from \cite{Skjetne2004mia}, and are presented in Table~\ref{tab: ship-Parameters}.
\begin{table}[htb]
	\centering
	\caption{3-DOF Cybership II model parameters}
	\label{tab: ship-Parameters}
	\begin{tabular}{lll | lll}
		\toprule[1.5pt]
		{\bf Par} & {\bf Value} & {\bf Unit} & {\bf Par} & {\bf Value} & {\bf Unit}\\
		\midrule
		$m$ & 23.8 & $kg$  & $Y_{|r|r}$& -0.02 &$\frac{kg}{m}$ \\
		$x_g$ & 0.046 & $m$    & $Y_{|v|r}$& -0.01 &$\frac{kg}{m}$ \\
		$I_z$ & 1.760 &$kg \cdot m^2$ & $Y_{|r|v}$ & -0.01 &$\frac{kg}{m}$\\
		$X_{u}$& -0.7225 &$\frac{N \cdot s}{m}$ & $N_v$ & 0.1052 &$\frac{kg \cdot s}{m}$ \\
		$X_{\dot{u}}$& -2.0 &$kg$ &	$N_{\dot{v}}$& 0.0 &$kg$\\
		$X_{|u|u}$& -1.3274 &$\frac{kg}{m}$ &	$N_{|v|v}$ & 5.0437 &$\frac{kg}{m}$ \\
		$X_{uuu}$& -5.8664 &$\frac{kg \cdot s}{m^2}$ &	$N_{r}$& -0.5 &$\frac{kg \cdot s}{m}$\\
		$Y_{v}$& -0.8612 &$\frac{N \cdot s}{m}$ & $N_{\dot{r}}$& -1.0 &$kg$\\
		$Y_{\dot{v}}$& -10.0 &$kg$ & $N_{|r|r}$& 0.005 &$\frac{kg}{m}$\\
		$Y_{|v|v}$& -36.2823 &$\frac{kg}{m}$& $N_{|v|r}$& -0.001 &$\frac{kg}{m}$\\
		$Y_{r}$& 0.1079 &$\frac{kg \cdot s}{m}$& $N_{|r|v}$& --0.001 &$\frac{kg}{m}$\\
            $Y_{\dot{r}}$& 0.0 &$kg$& & & \\

		\bottomrule
	\end{tabular}
        
\end{table}

\subsection{Disturbance modelling}
\label{sec:disturbance_modeling}
The sea current velocity $V_c$ and angle $\beta_c$ are modelled as slow-varying constrained random walk processes:
\begin{equation} \label{eq:disturbance_generation}
    \begin{aligned}
        \dot{V_{c}} & = W_{V_c}, \quad s.t. \; |V_{c}| \leq V_{c,max} \;, \; |W_{V_c}| \leq W_{V_c,max} \\
        \dot{\beta_{c}} & = W_{\beta_c}, \quad s.t. \; |\beta_{c}| \leq \beta_{c,max} \;, \; |W_{\beta_c}| \leq W_{\beta_c,max} \\
    \end{aligned}
\end{equation}

The generalized force disturbances are modelled similarly, with an additional white noise component added to the slowly varying signal:
\begin{equation}
    \begin{aligned} \label{eq:disturbance_forces}
        \force_d & = \bsym{\delta}_d + \bsym{w}_{\force 1}, \quad s.t. \; |\force_d| \leq \force_{d,max} \;, \; |\bsym{w}_{\force 1}| \leq \bsym{w}_{\force 1, max} \\
        \dot{\bsym{\delta}}_d & = \bsym{w}_{\force 2}, \quad s.t. \; |\bsym{w}_{\force 2}| \leq \bsym{w}_{\force 2, max}
    \end{aligned}
\end{equation}

To ensure a reasonable degree of controllability in the face of disturbances, the maximum current velocity $V_{c,max}$ is set to $\sim 20 \%$ of the maximum surge speed $u_{max}$. Similarly, the maximum surge and sway force disturbances are set to $\sim 20 \%$ of the maximum applied surge force $F_{u,max}$, while the maximum yaw moment disturbance is $\sim 10 \%$ of $T_{r,max}$

\subsection{Disturbance observer implementation}
\label{sec:disturbance_observer}

\begin{figure}[htb]
    \centering
    \begin{subfigure}{\linewidth}
        \includegraphics[width=\linewidth]{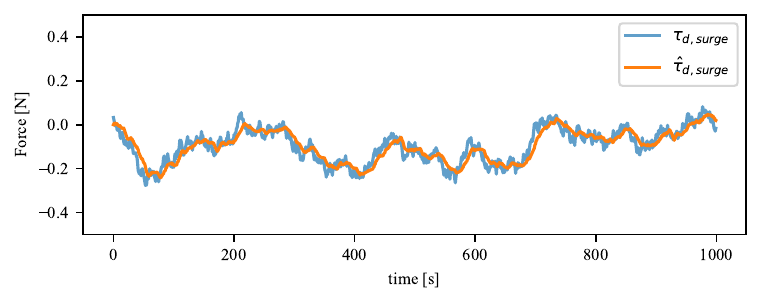}
        \caption{Estimate of surge force disturbance $\bsym{\tau}_{d, surge}$}
    \end{subfigure}
    \begin{subfigure}{\linewidth}
        \includegraphics[width=\linewidth]{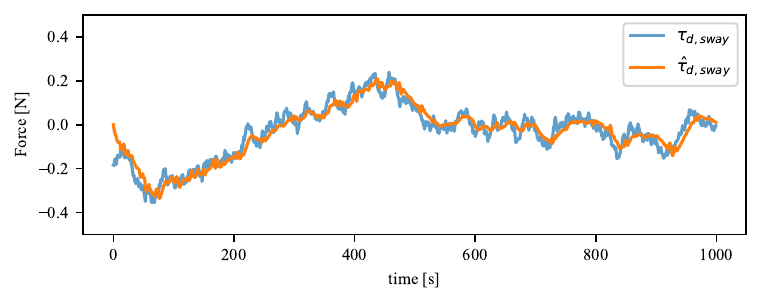}
        \caption{Estimate of sway force disturbance $\bsym{\tau}_{d, sway}$}
    \end{subfigure}
    \begin{subfigure}{\linewidth}
        \includegraphics[width=\linewidth]{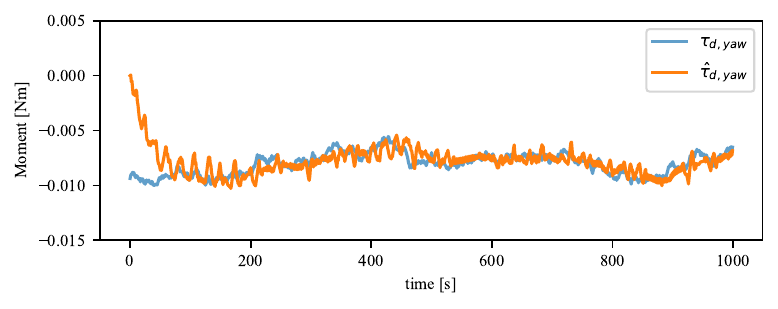}
        \caption[Environmental disturbance observer sample measurements]{Estimate of yaw moment disturbance $\bsym{\tau}_{d, yaw}$}
    \end{subfigure}
    \caption{Estimates of randomly generated environmental disturbances using adaption gains from Table~\ref{tab:disturbance_observer-params}}
    \label{fig:disturbance_estimation}
\end{figure}

The observer system \eqref{eq:observer_system_final} is implemented using a simple forward-euler scheme according to
\begin{equation}
    \begin{aligned}
        \hat{\force}_{d,k} & = \obsv_k + \bsym{T}\vel_k, \\
        \obsv_{k+1} & = \obsv_k - \bsym{T} \bsym{M}^{-1}(\bsym{D}(\vel_k)\vel_k - \bsym{C}(\vel_k)\vel_k + \bsym{B}\control_k + \hat{\force}_{d,k})\Delta T,
    \end{aligned}
\end{equation}
with the parameter values in Table~\ref{tab:disturbance_observer-params}.
The adaptation gains are chosen relatively small to ensure stability and sufficiently smooth disturbance estimates. Fig.~\ref{fig:disturbance_estimation} shows the typical performance of the estimator given randomized environmental disturbance forces generated according to \eqref{eq:disturbance_forces}.

\begin{table}[htb]
\centering
\caption{Environmental disturbance observer parameters}
\label{tab:disturbance_observer-params}
\begin{tabular}{c c c}
\hline
\textbf{Parameter} & \textbf{Description} & \textbf{Value} \\
\hline
$\Gamma_1$ & Adaptation gain (surge force) & 0.1 \\
$\Gamma_2$ & Adaptation gain (sway force) & 0.1\\
$\Gamma_3$ & Adaptation gain (yaw moment) & 0.08 \\
\bottomrule 
\end{tabular}
\end{table}

\subsection{PSF implementation}
\subsubsection{Collision avoidance}\label{sec:collision_avoidance_implementation}
In the 2-dimensional \textbf{gym-auv} simulation environment, static obstacles are represented as circles with a given position and radius, while other ships are represented as polygonal objects following pre-defined paths. In this work, we assume that the PSF has access to the position, shape, and velocity of ships that must be avoided. Additionally, the integrated LiDAR sensor suite is used to avoid static obstacle collisions.

\paragraph{Moving obstacle collision avoidance}
As stated above, we assume that the agent has access to the position, shape, and velocity of other moving ships at the current iterate. Each ship is given a circular hazard region centered at the midpoint of the ship, with a radius equal to the length of the ship. Denoting the position and velocity of ship $j$ at the current iterate as $\pos_{h,j}$ and $\boldsymbol{v}_{h,j}$ respectively, the moving obstacle collision avoidance constraint is implemented as
\begin{equation}
\begin{aligned}
    &||\pos_{i|k} - (\pos_{h,j} + \boldsymbol{v}_{h,j}i\Delta T)||_2 \geq l_{h,j} + d_{safe},\\ 
    &\quad \forall \; i \in [0,N], \; j \in [0,N_h], 
\end{aligned}  
\end{equation}
where $\Delta T$ is the step size, $l_{h,j}$ is the length of ship $j$ and $N_h$ is the number of ship obstacles. This formulation assumes that all ship obstacles move with constant velocities over the entire prediction horizon, which does not necessarily match the true trajectories of the obstacles. However, the results will show that satisfactory performance is achieved by choosing an appropriately large safety distance $d_{safe}$. 

\paragraph{LiDAR-based collision avoidance}

To encode collision avoidance constraints from the LiDAR measurements, the $N_{col} \leq \frac{N_{ray}}{N_{sector}}$ closest obstacle detections within each sector are extracted. For each of the extracted ray measurements, the coordinates of the point of detection $\pos_{detect}$ are calculated by
\begin{equation}
    \pos_{detect} = \bmat{x \\ y} + \bmat{d \cdot \cos{(\theta + \psi)} \\ d \cdot \sin{(\theta + \psi})},
\end{equation}
where $d$ is the distance to the object measured by the ray, and $\theta$ is the angle offset of the ray concerning the heading of the ship. The LiDAR collision avoidance constraint is then defined as
\begin{equation}
\begin{aligned}
    &||\pos_{i|k} - \pos_{detect,m}||_2 \geq R_{avoid} + d_{safe},\\ 
    &\quad \forall \; i \in [0,N], \; m \in [0,N_{col}N_{sector}],
\end{aligned}
\end{equation}
where $R_{avoid}$ defines the radius of avoidance around each detected point $\pos_{detect}$, and $N_{col}N_{sector}$ is the total number of included ray detections. As a result, the boundaries of nearby obstacles are approximated as a collection of circles centered at the detection points, with a radius equal to $R_{avoid}$. Safety is thus ensured by choosing a sufficiently high value for $R_{avoid}$ and having a sufficiently high number of $N_{col}$ detected points for each LiDAR sector. Fig.~\ref{fig:PSF_lidar_obstacles} shows how the LiDAR measurements are used to generate static obstacles.

\subsubsection{Terminal constraint computation}
The computation of the matrix $\bsym{P}_f$ which defines the terminal ellipsoidal invariant set and the corresponding control law $\bsym{k}_f(\state)$ follows the same procedure as used in \cite{Tearle2021aps} and \cite{Malmin2021rla}. The main idea is to linearize the system and constraints with regard to an equilibrium state, which allows us to construct a semidefinite program (SDP) that simultaneously optimizes the matrix $\bsym{P}_f$ and the terminal control law $\bsym{k}_f(\state)$.
We chose the equilibrium state equal to the maximum surge thrust and the corresponding constant maximum surge velocity, $\state_e = \bmat{u_{max} & 0 & 0 & 0 & 0 & 0}^T$ and $\control_e = \bmat{F_{u,max} & 0}^T$. The maximum surge velocity is found by solving for the steady state of the decoupled surge equation
\begin{equation}
    \begin{aligned}
        & 0 = -D_u(u_{max})u_{max} + F_{u,max}, \\
        & -X_uu_{max} - X_{|u|u}|u_{max}|u_{max} - X_{uuu}u_{max}^3 = F_{u,max}.
    \end{aligned}
\end{equation}

The next step is to linearize the state-space system concerning the equilibrium. To simplify the process, we omit the non-linear damping terms from the system model since the absolute value function $|\cdot |$ (which appears in $\bsym{D}(\vel)$) is non-differentiable at the origin. The non-linear damping is strictly dissipative wrt. surge velocity ($sign(-D_u(\vel)\vel) = -sign(u)$), and dissipative wrt. sway velocity unless $r >> v$, which is rarely the case in practice. The same does not hold for the yaw rate $r$ however, which means that the resulting control invariant set may have to be tightened to account for this. The simplified system model is given by
\begin{equation}
    \begin{aligned}
        \tilde{\state} & = \bmat{\tilde{\pose} & \tilde{\vel}}, \\
        \dot{\tilde{\pose}} & = \bsym{R}(\tilde{\psi})\tilde{\vel}, \\
        \dot{\tilde{\vel}} & = \bsym{M}^{-1}(-\bsym{C}(\tilde{\vel})\tilde{\vel} - \bsym{D}_L\tilde{\vel} + \control),
    \end{aligned}
\end{equation}
where $\bsym{D}_L$ is the linear damping
\begin{equation}
    \bsym{D}_L = \bmat{-X_u & 0 & 0 \\
                       0 & -Y_v & -Y_r \\
                       0 & -N_v & -N_r}.
\end{equation}

The linearized system is then computed as
\begin{equation}
    \dot{\bar{\state}} = \bsym{A}\bar{\state} + \bsym{B}\bar{\control},
\end{equation}
where
\begin{equation}
    \begin{aligned}
        \bar{\state} & = \tilde{\state} - \state_e, \\
        \bar{\control} & = \control - \control_e, \\
        \bsym{A} & = \bmat{\frac{\delta \dot{\tilde{\pose}}}{\delta \tilde{\pose}^T}|_{\state_e} & \frac{\delta \dot{\tilde{\pose}}}{\delta \tilde{\vel}^T}|_{\state_e} \\
        \bsym{0}^{3 \times 3} & \frac{\delta \dot{\tilde{\vel}}}{\delta \tilde{\vel}^T}|_{\state_e}}, \\
        \bsym{B} & = \bmat{\bsym{0}^{3 \times 2} \\
        \frac{\delta \dot{\tilde{\vel}}}{\delta \tilde{\control}^T}|_{\state_e, \control_e}}.
    \end{aligned}
\end{equation}
By defining the terminal control law $\bsym{k}_f(\state)$ as a linear feedback controller, applying it to the linearized system $\bsym{k}_f(\bar{\state}) = \bsym{K}\bar{\state}$, and using the closed-loop Lyapunov equation \cite{Tearle2021aps}, the constraint $\bar{\state}^T\bsym{P}_f\bar{\state} \leq 1$ can be rewritten as
\begin{equation}\label{eq:lyapunov}
    (\bsym{A + BK})^T\bsym{P}_f(\bsym{A + BK}) \prec 0.
\end{equation}

The final step before constructing the SDP is to transform the constraints into a linear polytopic form, expressed by
\begin{equation}\label{eq:polytopic}
    \begin{aligned}
        \bsym{H}\bar{\state} & \leq \bsym{h}, \\
        \bsym{G}\bar{\control} & \leq \bsym{g}.
    \end{aligned}
\end{equation}
Recall the definition of the terminal feasible set:
\begin{equation}
    \mathbb{C}_f := \{\state_f | \quad ||\pos_f - \pos_{N|k}||_2 \leq d_{f} \; \cap \; \vel_{lb} \leq \vel_f \leq \vel_{ub} \} 
    \tag{equation \eqref{eq:terminal-feasible-set} revisited}
\end{equation}
Without loss of generality, we assume $\pos_{N|k} = 0$, yielding
\begin{equation}
    \mathbb{C}_f := \{\state_f | \quad ||\pos_f||_2 \leq d_{f} \; \cap \; \vel_{lb} \leq \vel_f \leq \vel_{ub} \}.
\end{equation}
To satisfy the polytopic form requirement, we approximate the distance constraint $||\pos_f||_2 \leq d_{f}$ by imposing that $\pos_f$ must be inside the largest inscribed square of the circle with radius $d_f$ such that
\begin{equation} \label{eq:inscribed-square}
    \bmat{|x_f| \\ |y_f|} \leq \bmat{\frac{d_f}{\sqrt{2}} \\ \frac{d_f}{\sqrt{2}}} \rightarrow ||\pos_f||_2 \leq d_{f}.
\end{equation}
The construction of the polytopic constraints is now trivial, as the control input constraints and the remaining state constraints are simple bound constraints.

The largest constrained ellipsoidal set $\{\bar{\state}| \; \bar{\state}^T\bsym{P}_f\bar{\state} \leq 1\} \in \mathbb{C}_f$ can now be computed by solving the SDP (\cite{Tearle2021aps}, \cite{Malmin2021rla})
\begin{equation}\label{eq:semi_definite_program}
    \begin{aligned}
        \min_{\bsym{E}, \bsym{Y}} & -logdet(\bsym{E}), \\
        s.t. \quad & \bsym{E} \succeq 0, \\
        & \bmat{([\bsym{h}]_i - [\bsym{H}]_i\state_e)^2 & [\bsym{H}]_i\bsym{E} \\
                \bsym{E}[\bsym{H}]_i^T & \bsym{E}} \succeq 0 \quad \forall \; i \in [1, n_h], \\
        & \bmat{([\bsym{g}]_j - [\bsym{G}]_j\control_e)^2 & [\bsym{G}]_j\bsym{E} \\
                \bsym{E}[\bsym{G}]_j^T & \bsym{E}} \succeq 0 \quad \forall \; j \in [1, n_g], \\
        & \bmat{\bsym{E} & \bsym{EA}^T + \bsym{Y}^T\bsym{B}^T \\
                \bsym{AE} + \bsym{BY} & \bsym{E}} \succeq 0,
    \end{aligned}
\end{equation}
where $\bsym{E} = \bsym{P}_f^{-1}$ and $\bsym{Y} = \bsym{KE}$. From the computed optimal $\bsym{P}_f$ we can extract $\bsym{P}_{f_{\vel}}$. Note that $\bsym{P}_{f_{\vel}}$ and $\bsym{k}_f$ are only proven to be control invariant with respect to the linearized dynamics. A verification procedure should be applied, and $\bsym{P}_{f_{\vel}}$ iteratively refined until the invariance condition holds for the non-linear dynamics. For a detailed description of such a procedure, we refer the reader to \cite{Tearle2021aps}.  $\bsym{P}_{f_{\vel}}$ is inserted in the terminal velocity constraint $(i)$ of the predictive safety filter OCP (equation \eqref{eq:ship_PSF}).

\subsubsection{Predictive safety filter parameters}

From equation \eqref{eq:ship_PSF}, we define
\begin{equation}
    \control_{lb} = \bmat{F_{u,lb} \\ T_{r,lb}}, \quad \control_{ub} = \bmat{F_{u,ub} \\ T_{r,ub}}.
\end{equation}

The cost matrix $\boldsymbol{W}$ is given by
\begin{equation}
    \boldsymbol{W} = \bmat{\frac{\gamma_{F_u}}{(F_{u,ub} - F_{u,lb})^2} & 0 \\
                                 0 & \frac{\gamma_{T_r}}{(T_{r,ub} - T_{r,lb})^2}},
\end{equation}
where $\gamma_{F_u}$ and $\gamma_{T_r}$ are weighting constants, while the denominators ensure that the input signals are normalized according to their respective operating ranges.

The parameters used in the implementation of the predictive safety filter are shown in Table~\ref{tab:PSF-params}.

\begin{table}
\centering
\caption{Predictive safety filter parameters}
\label{tab:PSF-params}
\begin{tabular}{c c c}
\hline
\textbf{Parameter} & \textbf{Description} & \textbf{Value} \\
\hline
$N$ & Number of shooting nodes & 50 \\
$\Delta T$ & Discretization step  &  0.5 ($s$)\\
$F_{u,lb}$ & Minimum surge force  & -0.2 ($N$) \\
$F_{u,ub}$ & Maximum surge force  & 2 ($N$) \\
$T_{r,lb}$ & Minimum yaw moment & -0.15 ($Nm$) \\
$T_{r,ub}$ & Maximum yaw moment & 0.15 ($Nm$) \\
$N_{col}$ & Detection points per sector & 5 \\
$R_{avoid}$ & Detection point avoidance radius & 8 ($m$) \\
$d_{safe}$ & Minimum safe distance to hazards & 5 ($m$) \\
$\gamma_{F_u}$ & Surge force modification cost & 1 \\
$\gamma_{T_r}$ & Yaw moment modification cost & 0.01 \\
\end{tabular}
\end{table}

With 50 shooting nodes and a discretization step of 0.5$s$, the length of the prediction horizon is $T_f = 50 \cdot 0.5s = 25s$, which enables the safety filter to predict possible hazardous situations far in advance. Since $\gamma_{F_u} > \gamma_{T_r}$, the PSF is penalized less for applying modifications to the yaw moment (causing the ship to turn) as opposed to decreasing the surge force (causing the ship to slow down). As a consequence, the ship is more likely to steer away from potential hazards instead of slowing down to avoid them, which encourages forward progress.

The predictive safety filter is created with the \textbf{acados} nonlinear optimal control software \cite{Verschueren2022aam}, using a sequential-quadratic-programming real-time iteration scheme (SQP-RTI), and the internal QP-solver HPIPM \cite{Frison2020hah}. State constraints and collision avoidance constraints are implemented as soft constraints to guarantee feasibility. 

\subsection{Reward function}
The reward function employed is derived from the collision avoidance reward function proposed by Meyer \cite{Meyer2020ccc, Meyer2020taa}, and an additional term included for safety violations in the PSF is added. The reward function includes the main terms $r_{path}, r_{colav}$ and $r_{PSF}$, which provides rewards for path-following, collision avoidance, and safety violations. Additionally, there are two constant terms $r_{exists}$ and $r_{collision}$, which is the living and collision penalty. A more detailed explanation of the choice of the path and collision avoidance reward can be found in \cite{Meyer2020ccc, Meyer2020taa}. 

\subsubsection{Path reward}
The path reward has both a velocity-based and a CTE-based reward. The velocity-based reward is chosen to reward speed close to the maximum speed of the vessel $U_{max}$, while the heading term with the view, $\tilde{\psi}$, is small. The CTE-based reward penalizes large cross-track errors, with the expression going toward zero. To avoid the path reward being zero when the cross-track error is large, a constant term, $\gamma_r$, is included. 
\begin{equation}
    r_{\text {path }}^{(\mathrm{t})}=\underbrace{\left(\frac{u_{\max }^{(t)}}{U_{\max }} \cos \bar{\psi}^{(t)}+\gamma_r\right)}_{\text {Velocity-based reward }} \underbrace{\left(\exp \left(-\gamma_\epsilon\left|\epsilon^{(t)}\right|\right)+\gamma_r\right)}_{\text {CTE-based reward }}-\gamma_r^2
\end{equation}

\subsubsection{Collision avoidance reward}\label{sec:collision_avoidance_reward}
The collision avoidance reward penalizes the vessel for being close to obstacles heading towards it. It uses the LiDAR sensor measurements, where $d_i$ is the $i^{th}$ distance sensor measurement, $\theta_i$ is the vessel-relative angle of the corresponding sensor ray, and $v_{y}^{i}$ is the y-component of the $i^{th}$ velocity measurement. The final expression accounts for both static and dynamic obstacles and is the following weighted average.
\begin{equation}
\begin{gathered}
r_{\text {colav }}^{(t)}=-\frac{\sum_{i=1}^N \frac{1}{1+\gamma_\theta\left|\theta_i\right|} \exp \left(\gamma_v \max \left(0, v_y^i\right)-\gamma_x d_i\right)}{\sum_{i=1}^N \frac{1}{1+\gamma_\theta\left|\theta_i\right|}}
\end{gathered}
\end{equation}

\subsubsection{Safety violation reward}\label{sec:PSF_penalty}
In order to avoid that the agent relies on the PSF, a negative reward is added for actions that violate the constraints in the PSF. The weighting factor $\gamma_{PSF}$ decides how much the agent is penalized for being corrected by the PSF.
\begin{equation}
    \begin{aligned}
r_{P S F}^{(t)} & =-\gamma_{P S F}| \frac{u_{\mathcal{L}}-u_0}{u_{\max }} | \\
& =-\gamma_{P S F}\left(\left|\frac{F_{u}-F_{u, P S F}}{F_{u, \max }}\right|+\left|\frac{T_r-T_{r, P S F}}{T_{r, \max }}\right|\right)
\end{aligned}
\end{equation}

\subsubsection{Complete reward function}
The final expression includes the two constant terms for living and collision penalty, $r_{exists}$ and $r_{collision}$. Additionally, the parameter $\lambda$ is provided to regulate the trade-off between the path and collision avoidance reward. 
\begin{equation}
r= \begin{cases}r_{\text {collision }}, & \text { if collision } \\
\lambda r_{\text {path }}+(1 - \lambda) r_{\text {colav }}+ r_{P S F}+ r_{\text {exists }}, & \text { otherwise }\end{cases}
\end{equation}

\FloatBarrier

\subsection{Test Cases}
To assess the performance of the PPO + PSF agent, four test cases were defined with different levels of difficulty and complexity. For test cases 1 and 2 (Fig.~\ref{fig:case1_and_2_examples}), disturbances are not included. Removing disturbances allow us to verify the implementation of the PSF, and study the impact of the PSF on the learning-rate and behavior of the agent in a controlled setting. For test cases 3 and 4, environmental disturbances are added to increase the difficulty and realism of the training and test scenarios. In the training environment, each episode was run until one of three termination criteria was satisfied: the distance to the goal location was less than 5 meters, the path progress exceeded 99\%, or the maximum number of timesteps for the specific case was reached. A description of each test case follows. 

\subsubsection{Case 1: Predescribed path with stationary obstacles}
\begin{figure}[h]
    \centering
    \begin{subfigure}[h]{0.45\linewidth}
        \centering
        \includegraphics[width=\linewidth]{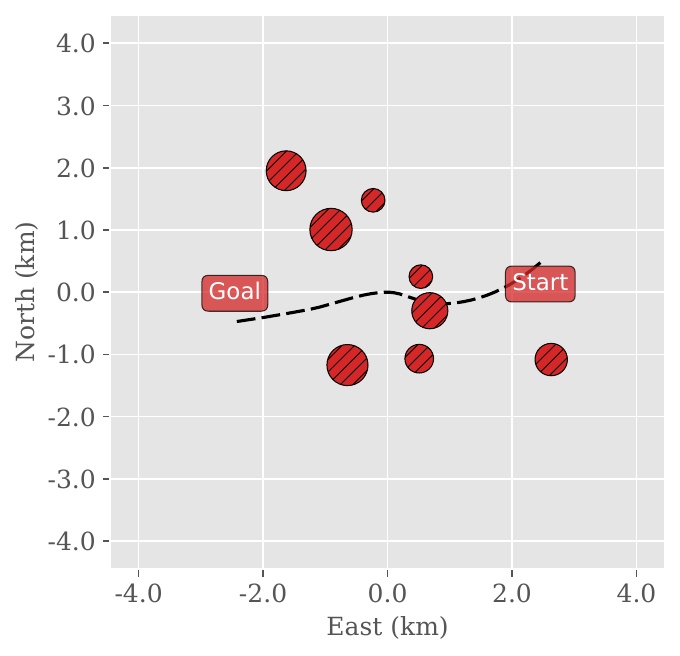}
        \caption{Scenario generated in case 1}
        \label{fig:case1_example}
    \end{subfigure}
    \begin{subfigure}[h]{0.45\linewidth}
        \centering
        \includegraphics[width=\linewidth]{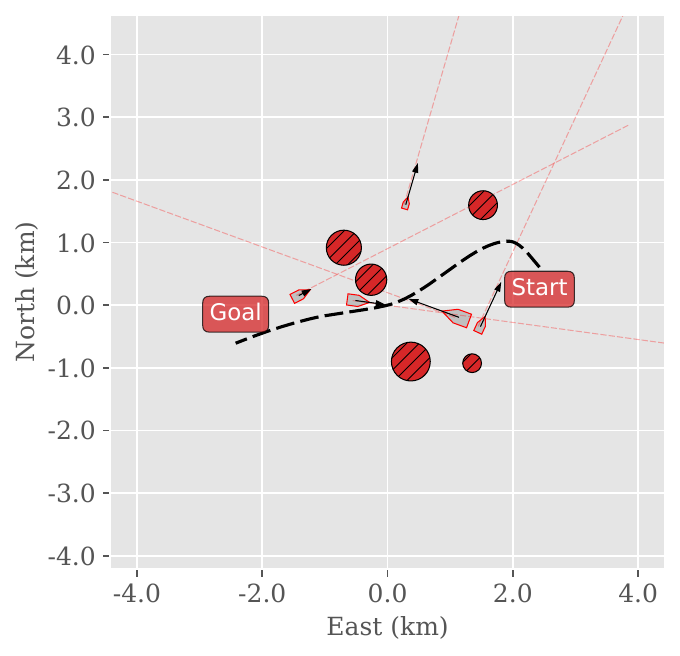}
        \caption{Scenario generated in case 2}
        \label{fig:case2_example}
    \end{subfigure}
    \caption[Randomly generated scenarios for test cases 1 and 2]{Sample of two randomly generated scenarios for test cases 1 and 2. Red circles indicate static obstacles, which are randomized both in terms of position and size. Dotted black curve indicates path, which might be obstructed by obstacles, in which case the agent should find the minimum necessary deviation from the path. In the right-hand figure, red polygons indicate target ships, while the red lines show their trajectories. Target ships are allowed to pass through static obstacles and each other, because otherwise, prohibitively complex randomization procedures would be necessary.}
    \label{fig:case1_and_2_examples}
\end{figure}

The first case considers a randomly generated path with stationary obstacles. Each obstacle is generated with random size and position according to the pre-set parameters $\mu_{\text {r, stat }}$ and $\sigma_d$. Table~\ref{tab:case-1-parameters} shows the chosen parameters for the scenario and Fig.~\ref{fig:case1_example} depics an example of a generated scenario. This scenario is a good starting point to evaluate how the inclusion of the PSF changes the performance in a basic obstacle avoidance setting compared to the standard PPO algorithm, which already has demonstrated good results in similar scenarios \cite{Meyer2020ccc, Meyer2020taa}. 

\begin{table}
\centering
\caption{Parameters for generating case 1}
\label{tab:case-1-parameters}
\begin{tabular}{c c c}
\hline
\textbf{Parameter} & \textbf{Description} & \textbf{Value} \\
\hline
$N_{o, stat}$ & Number of static obstacles & 8 \\
$N_{w}$ & Number of path waypoints  &  2\\
$L_{p}$ & Path length  & 500 \\
$\mu_{\text {r, stat }}$ & Mean static obstacle radius & 30 \\
$\sigma_d$ & Obstacle displacement std. dev. & 100 \\
\end{tabular}
\end{table}

\subsubsection{Case 2: Predescribed path with stationary and moving obstacles}
The second case includes moving obstacles to simulate ships in a real marine environment. The moving obstacles are spawned similar to the static obstacles and follow linear trajectories. Table~\ref{tab:case-2-parameters} shows the chosen parameters and Fig.~\ref{fig:case2_example} shows an example of a generated scenario. 

\begin{table}[h]
\centering
\caption{Parameters for generating case 2}
\label{tab:case-2-parameters}
\begin{tabular}{c c c}
\hline
\textbf{Parameter} & \textbf{Description} & \textbf{Value} \\
\hline
$N_{o, stat}$ & Number of static obstacles & 5 \\
$N_{o, dyn}$ & Number of dynamic obstacles & 5 \\

$N_{w}$ & Number of path waypoints  & 2 \\
$L_{p}$ & Path length  & 500 \\
$\mu_{\text {r, stat }}$ & Mean static obstacle radius & 25 \\
$\mu_{\text {r, dyn }}$ & Mean moving obstacle radius & 15 \\
$\sigma_d$ & Obstacle displacement std. dev. & 100 \\
\end{tabular}
\end{table}

\subsubsection{Case 3: Predescribed path with stationary and moving obstacles and disturbances}
In test case 3, randomized environmental disturbances are added to the simulations in order to assess the robustness of the predictive safety filter under more realistic and challenging conditions. The ocean current and disturbance forces are generated according to equation \eqref{eq:disturbance_generation}. Both static and dynamic obstacles are included, and randomly generated using the same parameters as in Case 2. The environmental disturbances are estimated using the observer described in Section \ref{sec:disturbance_observer}, and these estimates are consequently included in both the predictive safety filter and the observation vector of the reinforcement learning agent. 
    
\subsubsection{Case 4: Real environment}
\begin{figure}[htbp]
  \centering
  \begin{subfigure}[b]{0.32\linewidth}
    \includegraphics[width=\linewidth]{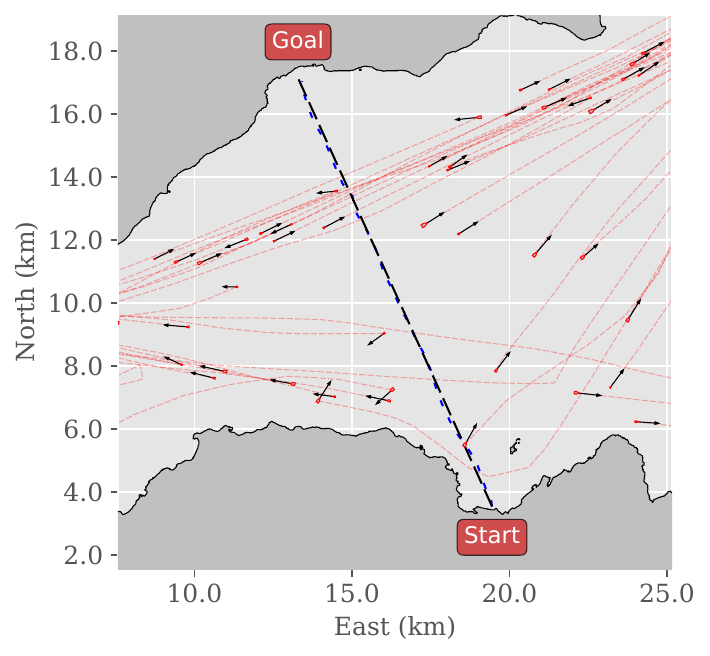}
    \caption{Trondheim}
    \label{subfig:Trondheim_path}
  \end{subfigure}
  \begin{subfigure}[b]{0.32\linewidth}
    \includegraphics[width=\linewidth]{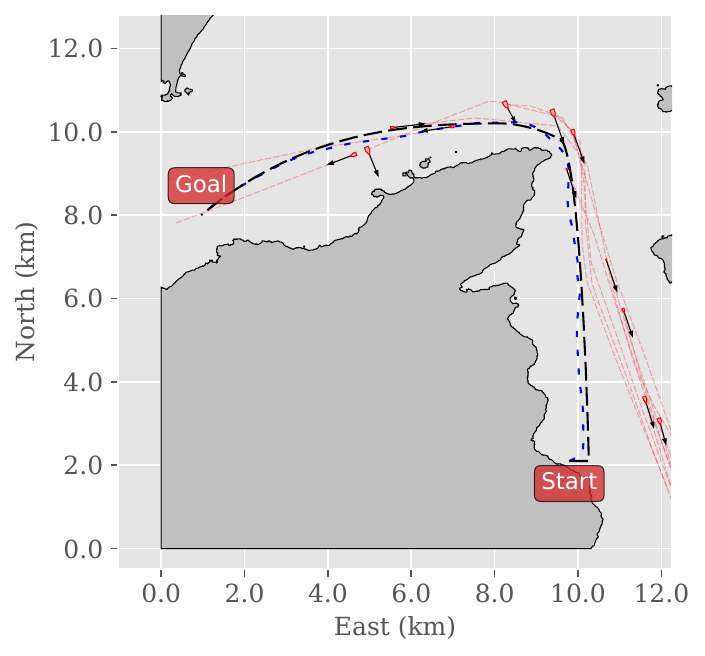}
    \caption{Agdenes}
    \label{subfig:Agdenes_path}
  \end{subfigure}
  \begin{subfigure}[b]{0.32\linewidth}
  \centering
    \includegraphics[width=\linewidth]{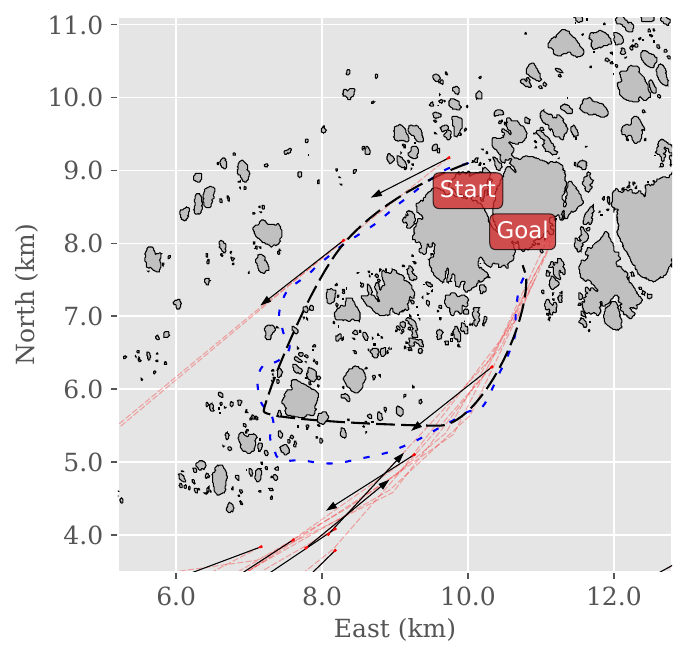}
    \caption{Sørbuøya}
    \label{subfig:Sorbuoya_path}
  \end{subfigure}
  \caption{PPO+PSF agent trajectories in the real-world scenarios. The blue dotted line is the path taken by agent and the red dotted lines are the path taken by the moving obstacles. The agent stays close to the desired path in all scenarios except for Sørbuøya, where it has to deviate in order to avoid collision with islets}
  \label{fig:real_world_scenarios_results}
\end{figure}

Finally, the fourth case evaluates the algorithm's performance in more realistic marine environments. These environments were developed by Meyer \cite{Meyer2020taa, Meyer2020ccc} and include terrain data from the Trondheim fjord and AIS tracking data from vessels in the area. There are in total three challenging environments that require a different set of skills to navigate. The \textbf{Trondheim} scenario requires the vessel to follow a straight path to cross the fjord while avoiding traffic from multiple crossing ships. In the \textbf{Agdenes} scenario, the vessel has to blend in with two-way traffic in order to avoid collisions in a narrow area at the entrance of the Trondheimsfjord. Lastly, for the \textbf{Sørbuøya} scenario, the vessel has to navigate through hundreds of small islands to reach the goal, which requires proficient static obstacle avoidance. Each scenario is generated with a random sample of ships from an AIS database, such that the vessel will face a variety of different traffic situations. The disturbances from Case 3 were also included in this scenario, to further improve the realism. A snapshot of the different events with the trajectories of the ships can be seen in Fig.~\ref{fig:real_world_scenarios_results} illustrating the path taken by the RL agent in the three real-world scenarios.

\subsection{Evaluation}
The evaluation aims to compare the standard PPO algorithm with the PPO + PSF algorithm to demonstrate how inclusion of the PSF changes the learning process and agent behavior. The assessment of the agent's performance was twofold. First, performance during training in the environment was evaluated. The test results were then examined to compare the final performance of the two agents.

\subsubsection{Training performance.}
To evaluate the performance of the training, we first consider the number of collisions during the training. The standard PPO agent is expected to have several collisions early on, but later go towards zero when a good policy is learned. On the other hand, the PPO + PSF agent is expected to have zero collisions during the entire training process, since the PSF should be able to correct the actions of any suboptimal policy in order to avoid collisions. Therefore, by comparing the number of collisions, we should be able to differentiate the two algorithms. It is also a good indicator that the PSF ensures safe behavior. Additionally, statistics on reward and cross-track error during training are used to evaluate overall performance and understand how the inclusion of a PSF influences the learning process.

\subsubsection{Test performance.}
The trained agents are tested in 100 randomly generated scenarios to evaluate their final performance. This sample size was chosen as it provides statistically significant results, while keeping computational time manageable. We decided to use slightly different performance metrics for the test results. While the average reward and cross-track error are useful for observing the learning progress and overall performance in terms of reward, these metrics become less relevant in a practical setting. More critical is whether the agent reaches the goal within a reasonable time and without collisions. Therefore, we used the average path progress and time consumption as performance indicators, in addition to collision avoidance. While the path progress and collision avoidance are straightforward to calculate, the time consumption metric requires some explanation. The minimum possible time was set as the path length divided by the maximum speed of the vessel, which corresponds to 100\%. Note that this is usually not possible to achieve in practice. The maximum time was defined as the maximum number of time steps allowed in a scenario before the episode was terminated, which was set to 5000 seconds in all cases, corresponding to 0\%. All episodes with a collision were excluded from the time consumption average.

In addition to statistical data collected during training and testing, figures illustrating the agents' behavior in various scenarios are also provided. These can assist in making more qualitative assessments of the performance.

\section{Results and discussions}
\label{sec:resultsanddiscussions} 
\subsection{Training results}
\subsubsection{Case 1: Predescribed path with stationary obstacles}

Each agent was trained for 1 million timesteps, which corresponded to around 850 episodes. The exact number depended on the time spent to complete each scenario. 

\begin{figure}[h]
\centering
\begin{subfigure}[h]{0.32\linewidth}
\centering
\adjincludegraphics[width=\textwidth, trim={0.021\width, 0.027\height, 0.096\width, 0.118\height},clip]{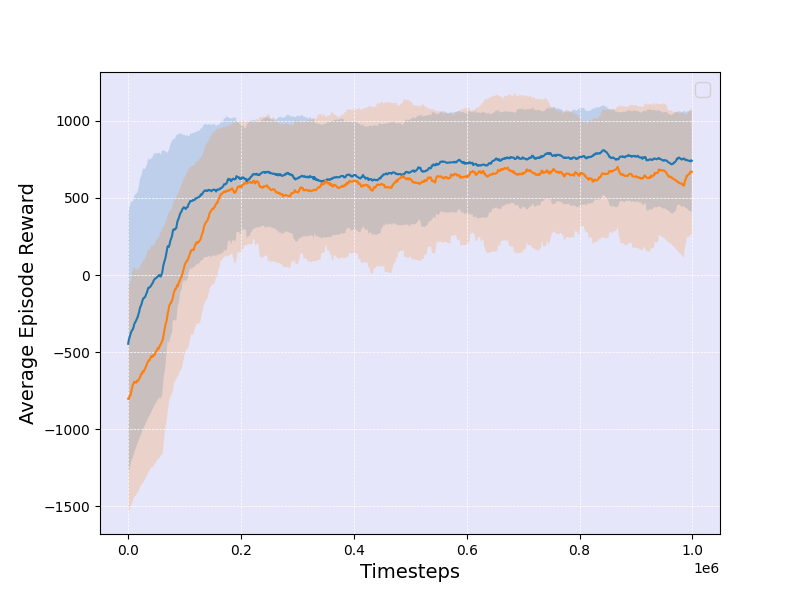}
\subcaption{Reward}
\end{subfigure}
\begin{subfigure}[h]{0.32\linewidth}
\centering
\adjincludegraphics[width=\textwidth, trim={0.021\width, 0.027\height, 0.096\width, 0.118\height},clip]{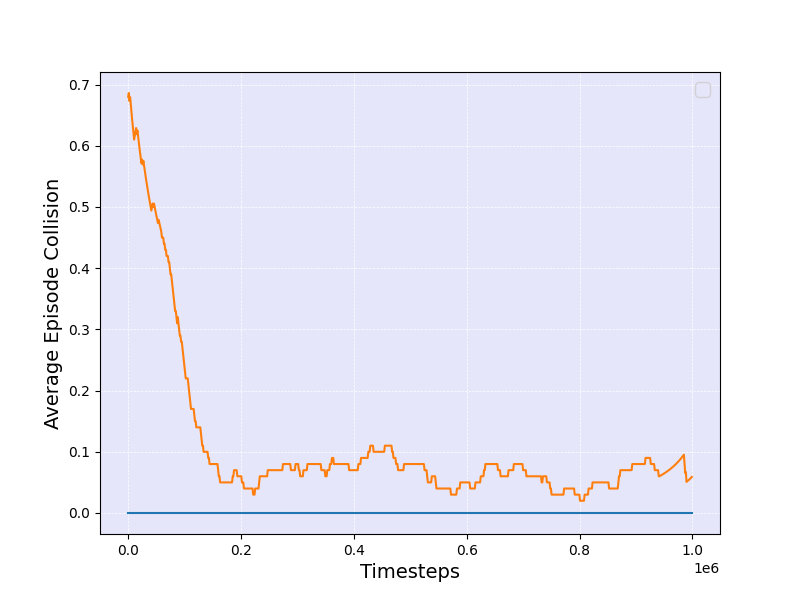}
\subcaption{Collision rate}
\end{subfigure}
\begin{subfigure}[h]{0.32\linewidth}
\centering
\adjincludegraphics[width=\textwidth, trim={0.021\width, 0.027\height, 0.096\width, 0.118\height},clip]{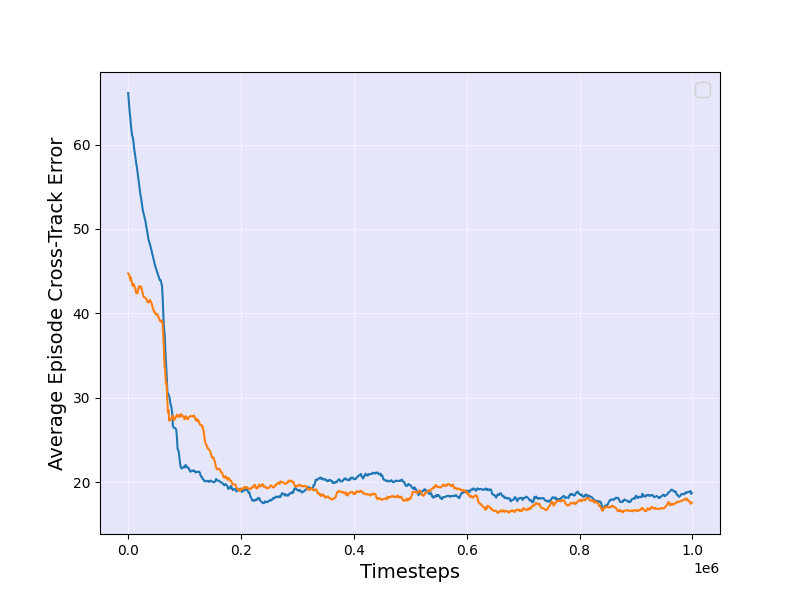}
\subcaption{Cross-Track Error}
\label{fig:case1_training_CTE}
\end{subfigure}
\begin{subfigure}[h]{\textwidth}
\centering
\includegraphics[width=0.3\textwidth]{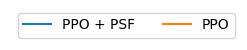}
\end{subfigure}
\caption{Average reward, collisions, and cross-track error during training smoothed with a rolling average over 100 episodes in Case 1. The collision rate is zero for the PPO+PSF agent during the entire training period. The difference in collision rate is especially striking during the first 200,000 timesteps of training, where the standard PPO agent crashes in approximately 40\% of the episodes}
\label{fig:case1_training}
\end{figure}

As can be seen in Fig.~\ref{fig:case1_training}, no collisions occurred for the PPO+PSF agent in case 1. The PPO agent conversely experienced a high collision rate early on, but eventually reached a rate near zero. The reward for the PPO+PSF agent is slightly higher throughout the training, mostly due to the absence of collisions. Additionally, the cross-track error is slightly lower for the PPO agent, which is expected since the PSF requires the agent to be at a minimum constant distance to every obstacle, which in some cases would mean that it has to stay further away from the path. 

\begin{figure}[h]
  \centering
  \begin{subfigure}[b]{0.45\linewidth}
    \includegraphics[width=\textwidth]{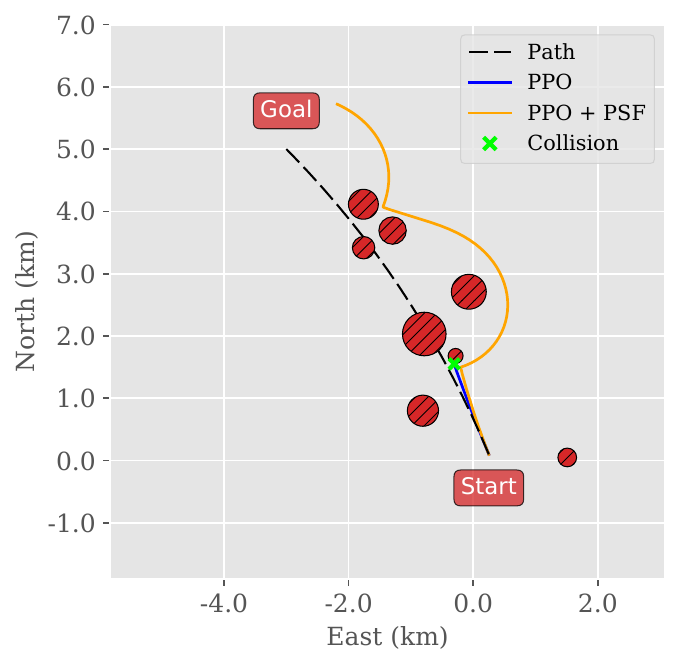}
    \caption{Agent trained for 10.000 timesteps.}
    \label{fig:10000_ts}
  \end{subfigure}
  \hfill
  \begin{subfigure}[b]{0.45\linewidth}
    \includegraphics[width=\textwidth]{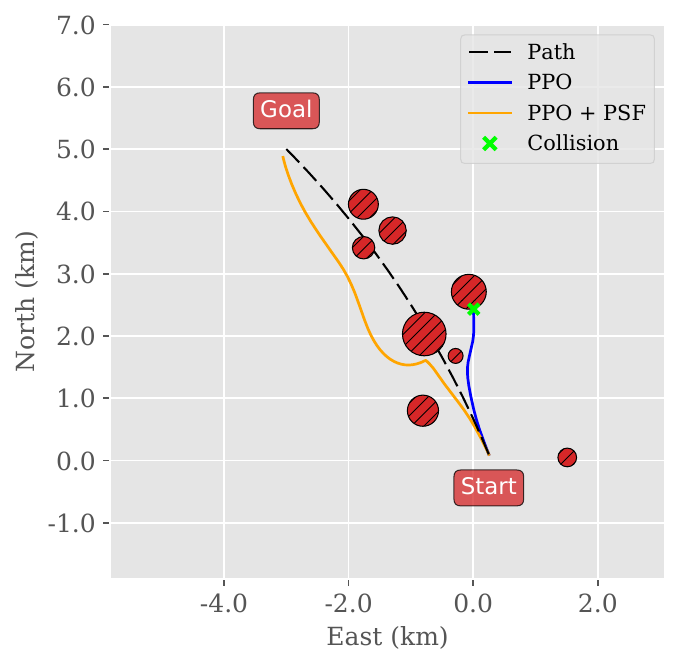}
    \caption{Agent trained for 50.000 timesteps}
    \label{fig:50000_ts}
  \end{subfigure}

  \begin{subfigure}[b]{0.45\linewidth}
    \includegraphics[width=\textwidth]{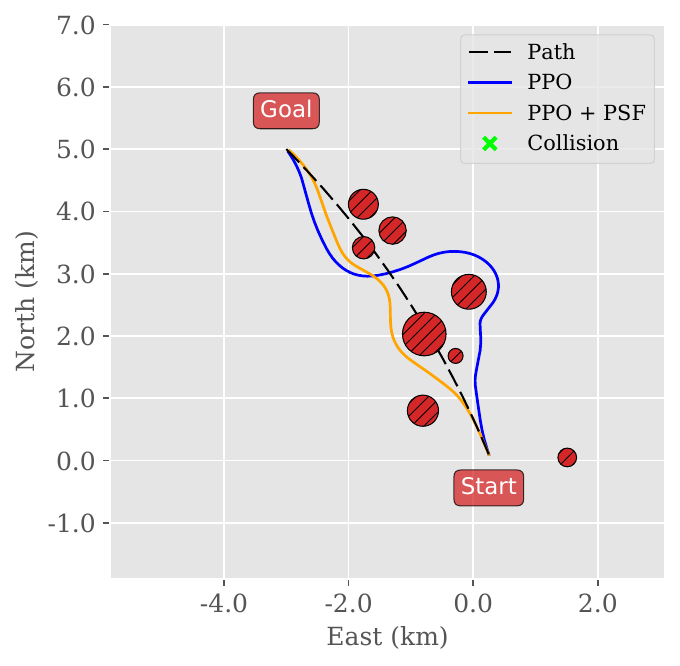}
    \caption{Agent trained for 100.000 timesteps.}
    \label{fig:100000_ts}
  \end{subfigure}
  \hfill
  \begin{subfigure}[b]{0.45\linewidth}
    \includegraphics[width=\textwidth]{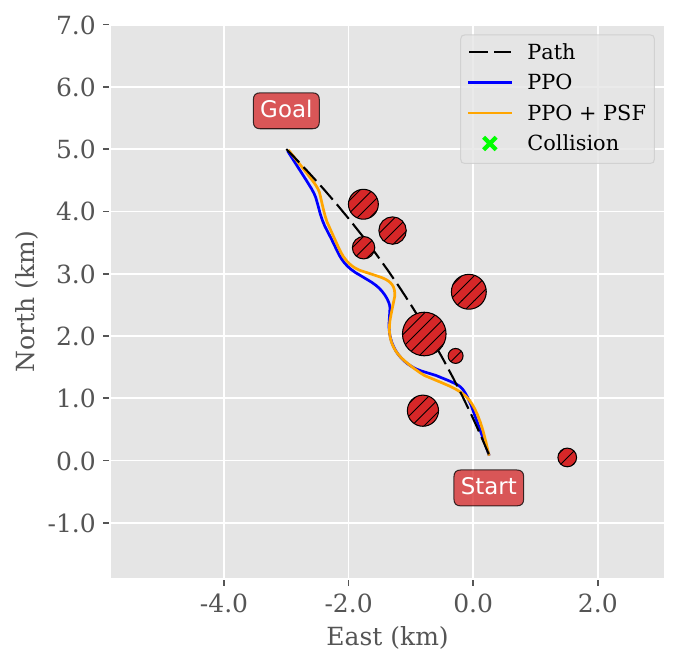}
    \caption{Agent trained for 400.000 timesteps.}
    \label{fig:400000_ts}
  \end{subfigure}
  \caption{Comparison between the PPO and the PPO+PSF agent at different stages of training. Notice that the PPO+PSF agent converges faster to an optimal trajectory for this specific scenario.}
  \label{fig:case1_agent_comparison}
\end{figure}

A comparison between the agent's trajectories at different stages of training can be seen in Fig.~\ref{fig:case1_agent_comparison}. 
These agents were trained in random scenarios in case 1 for different durations and tested in a sample scenario to compare their performance. After 10.000 timesteps, the agent performs quite poorly, with the PPO agent crashing at an early stage. The PSF saves the PPO+PSF agent from crashing similarly by modifying the action to perform a sharp right turn before the obstacle is reached. Notice that even though the agent does not reach the goal exactly, the scenario finishes since the path progress is above 99 \%, which is one of the termination conditions. Gradually, as the agents learn, we observe that the PPO agent becomes better at collision avoidance and that both agents achieve a lower cross-track error. After 400.000 timesteps both agents follow what seems to be a close to optimal trajectory for this specific scenario. Interestingly, the PPO+PSF agent seems to converge to the optimal trajectory faster than the PPO agent. While this is less evident in the cross-track error plot in Fig.~\ref{fig:case1_training_CTE}, it could suggest that the absence of collisions early on in the training accelerates the learning process. 

\begin{figure}
\centering
\begin{subfigure}[h]{0.45\linewidth}
\centering
\includegraphics[width=\textwidth]{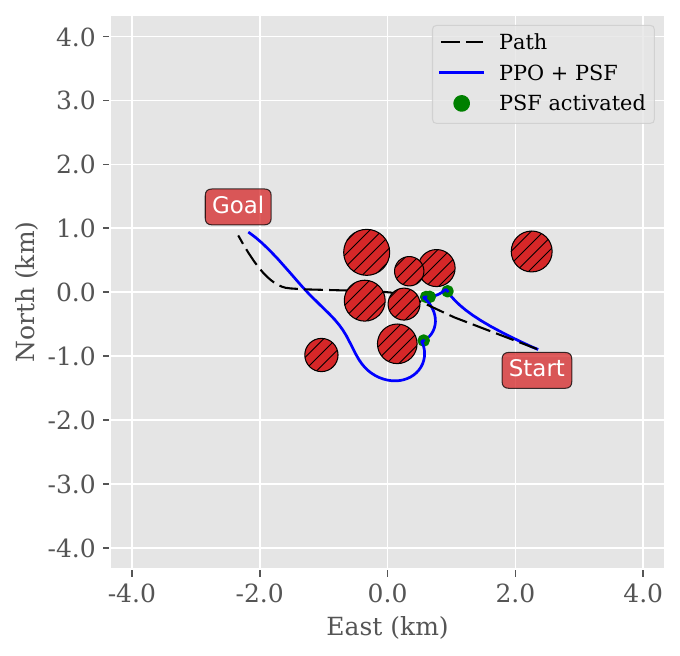}
\subcaption{Scenario 1}
\end{subfigure}
\begin{subfigure}[h]{0.45\linewidth}
\centering
\includegraphics[width=\textwidth]{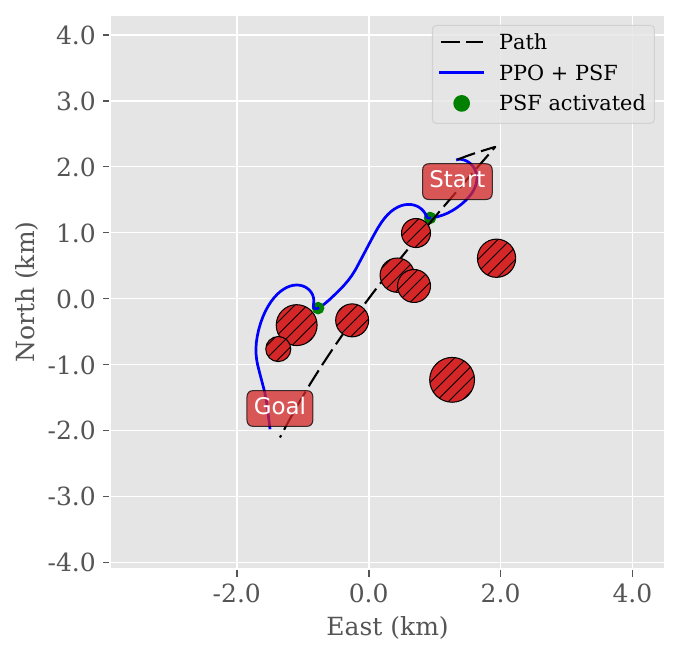}
\subcaption{Scenario 2}
\end{subfigure}
\caption{PSF corrections in two scenarios for an agent trained for 50.000 timesteps. The PSF prevents a collision multiple times to ensure that the agent reaches the goal destination. Because the PSF only optimizes with respect to finding the minimum perturbation to the proposed agent input, the PSF does not modify the control input until the last timestep before a collision is unavoidable. Therefore the agent is very close to the obstacles before control input modifications occur.}
\label{fig:case1_PSF_corrections}
\end{figure}

Furthermore, Fig.~\ref{fig:case1_PSF_corrections} illustrates instances when the PSF is activated during training. The green dots visualize where the agents' intended action is regarded as unsafe, and therefore corrected by the PSF. Because the PSF avoids situations that otherwise would become collisions, the episode length for the PPO+PSF agent is significantly higher in the early stages of training compared with the pure PPO agent. Additionally, within a single episode, the agent can encounter multiple situations that would have resulted in crashes without the PSF. This enables the agent to experience more unsafe states within less time. As a result, the PPO+PSF agent is able to learn more from fewer training episodes.

\subsubsection{Case 2: Predescribed path with stationary and moving obstacles}

\begin{figure}[h!]
\centering
\begin{subfigure}[h]{0.32\linewidth}
\centering
\adjincludegraphics[width=\textwidth, trim={0.021\width, 0.027\height, 0.096\width, 0.118\height},clip]{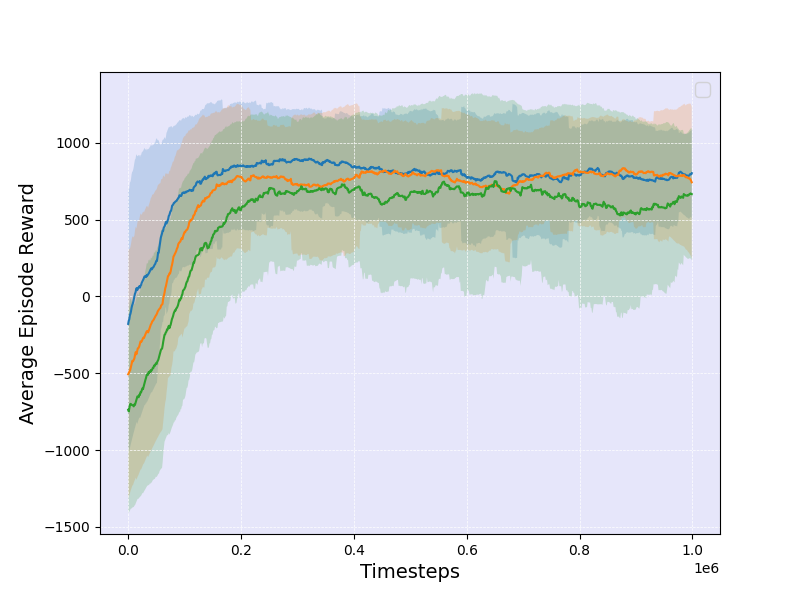}
\subcaption{Reward}
\label{fig:case2_reward}
\end{subfigure}
\begin{subfigure}[h]{0.32\linewidth}
\centering
\adjincludegraphics[width=\textwidth, trim={0.021\width, 0.027\height, 0.096\width, 0.118\height},clip]{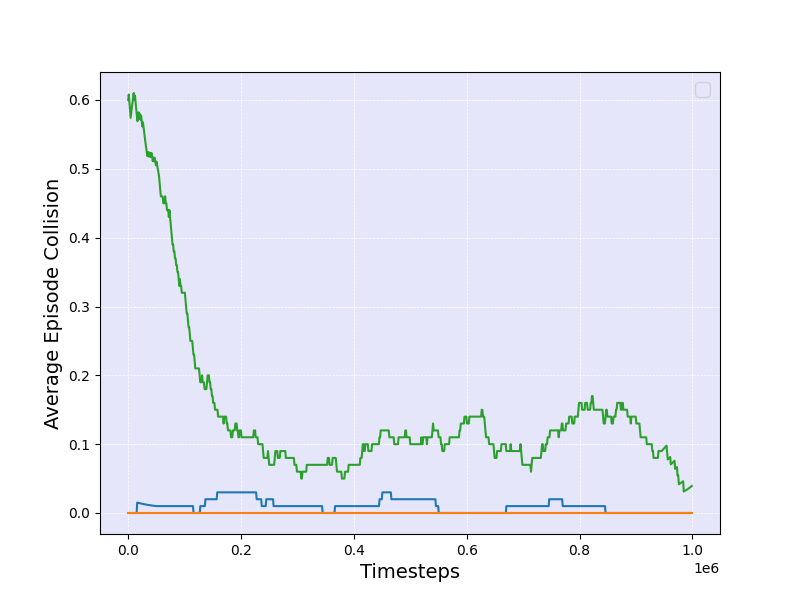}
\subcaption{Collision rate}
\label{fig:case2_collision}
\end{subfigure}
\begin{subfigure}[h]{0.32\linewidth}
\centering
\adjincludegraphics[width=\textwidth, trim={0.021\width, 0.027\height, 0.096\width, 0.118\height},clip]{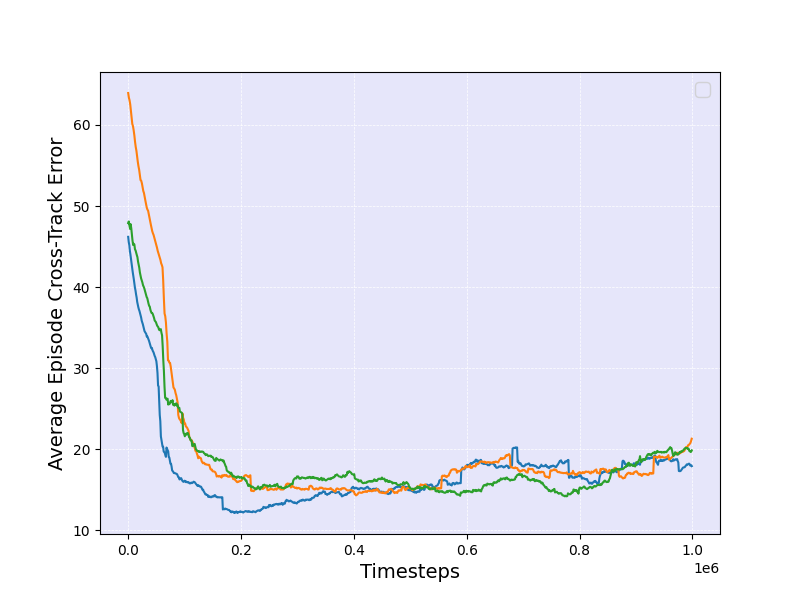}
\subcaption{Cross-track error}
\label{fig:case2_cte}
\end{subfigure}\\
\begin{subfigure}[h]{\linewidth}
\centering
\includegraphics[width=\linewidth]{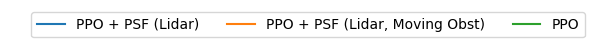}
\end{subfigure}
\caption{Average reward, collisions, and cross-track error during training smoothed with a rolling average over 100 episodes in Case 2. Only the PPO+PSF with information about the moving obstacles has a collision rate of zero during the entire training period.}
\label{fig:case2_training}
\end{figure}

In Case 2 we observe that the PSF that uses only LiDAR for collision avoidance no longer managed to prevent all collisions during training, as shown in Fig.~\ref{fig:case2_training}. However, providing the PSF access to the position and velocity of nearby target ships (encoded via moving obstacle constraint as defined in section \ref{sec:collision_avoidance_implementation}) leads to better decisions for avoiding unsafe situations. The approach proved to be effective, resulting in zero collisions during both training and testing. The standard PPO agent struggled more in this case, with a higher collision rate than in the previous case, especially for the first 200,000 timesteps, which can be seen in Fig.~\ref{fig:case2_collision}. For scenarios including moving obstacles, the PSF vs. PPO comparison is slightly unfair due to the difference in available information; the baseline agent does not have access to the obstacle velocities. Theoretically, we could simply append this information to the agent's observation vector. However, this would, in turn, result in a significant increase in the state space for the agent to explore in the training phase. A considerable increase in the input will likely lead to longer training times and can inhibit convergence toward a sufficient policy. Finding an efficient representation that minimizes the increase in state space while coupling the velocities to the corresponding obstacles in the LiDAR sensor is left as future work.

\subsubsection{Case 3: Predescribed path with stationary and moving obstacles and
disturbances}

\begin{figure}[h!]
\centering
\begin{subfigure}[h]{0.32\linewidth}
\centering
\adjincludegraphics[width=\textwidth, trim={0.021\width, 0.027\height, 0.096\width, 0.118\height},clip]{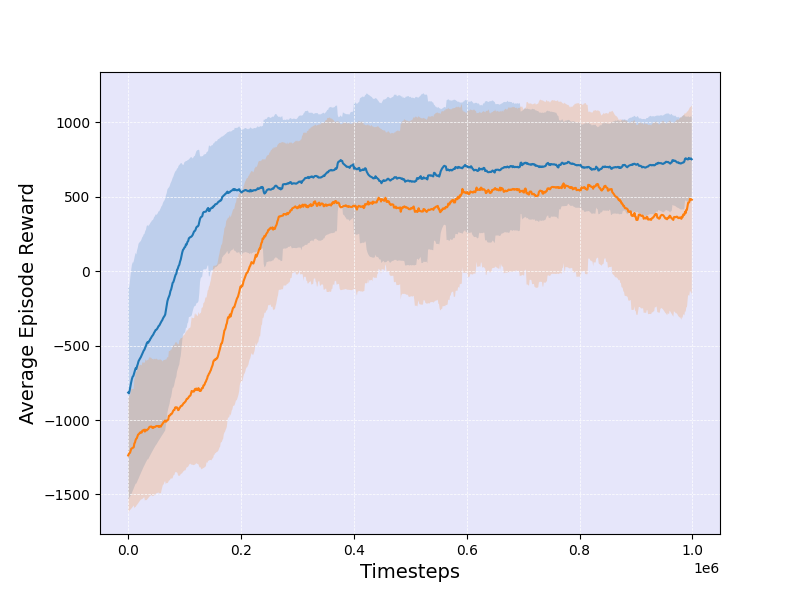}
\subcaption{Reward}
\label{fig:case3_reward}
\end{subfigure}
\begin{subfigure}[h]{0.32\linewidth}
\centering
\adjincludegraphics[width=\textwidth, trim={0.021\width, 0.027\height, 0.096\width, 0.118\height},clip]{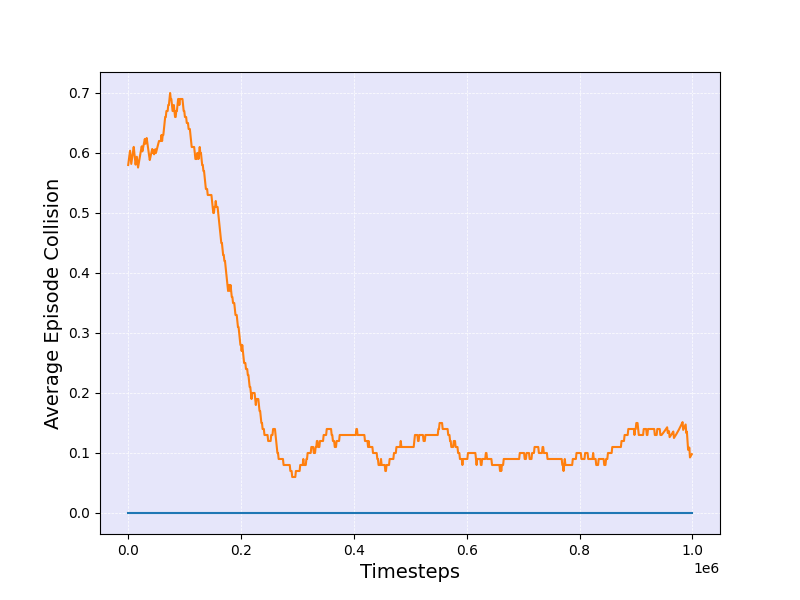}
\subcaption{Collision rate}
\label{fig:case3_collision}
\end{subfigure}
\begin{subfigure}[h]{0.32\linewidth}
\centering
\adjincludegraphics[width=\textwidth, trim={0.021\width, 0.027\height, 0.096\width, 0.118\height},clip]{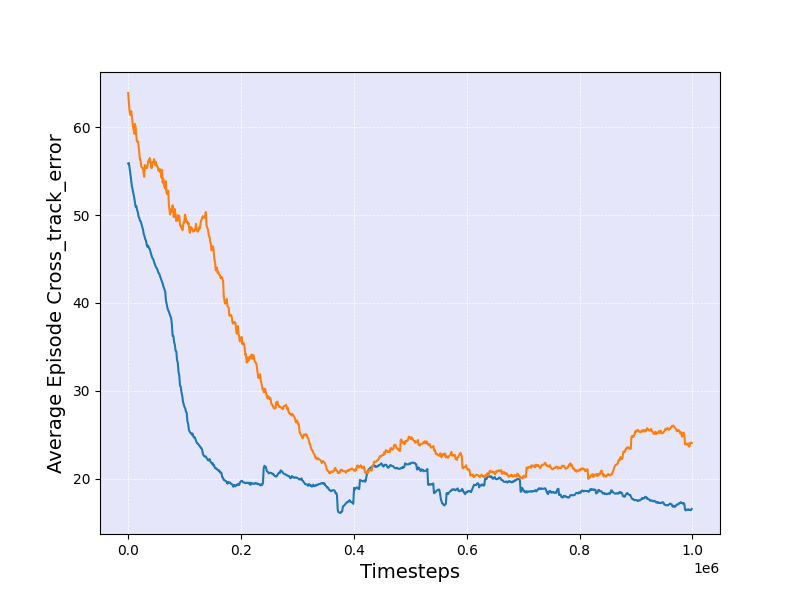}
\subcaption{Cross-Track Error}
\label{fig:case3_cte}
\end{subfigure}\\
\begin{subfigure}[h]{\linewidth}
\centering
\includegraphics[width=\linewidth]{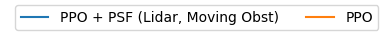}
\end{subfigure}
\caption{Average reward, collisions, and cross-track error during training, smoothed with a rolling average over 100 episodes in Case 3. The added disturbances lower the overall performance of the agents, but they still converge to a satisfactory level}
\label{fig:case3_training}
\end{figure}

In this case, the observation vector was augmented with the disturbance estimates described in Section \ref{sec:disturbance_observer}, to account for the added disturbances. The training results in Fig.~\ref{fig:case3_training} show that the PPO agent took longer to reach a low collision rate, compared to previous cases. Both the cross-track error and the total reward were slightly lower in this case, as expected. However, the collision rate for the standard PPO agent is still noticeably high during the initial 200,000 timesteps of training, as seen in Fig.~\ref{fig:case3_collision}.

\subsection{Test results}

\begin{figure}[h!]
\centering
  \begin{subfigure}[b]{0.32\linewidth}
\includegraphics[width=\textwidth]{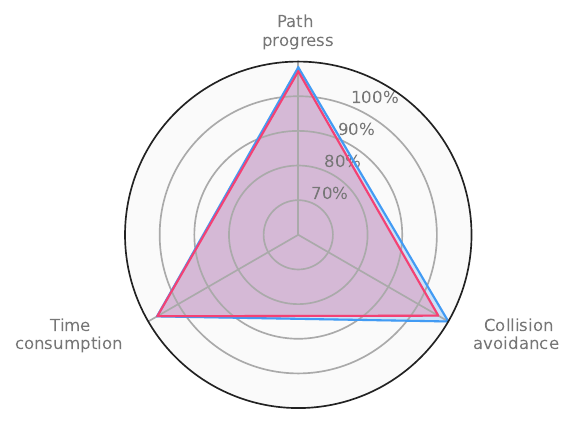}
\caption{Case 1}
\label{fig:case1_test}
  \end{subfigure}
  \begin{subfigure}[b]{0.32\linewidth}
\includegraphics[width=\textwidth]{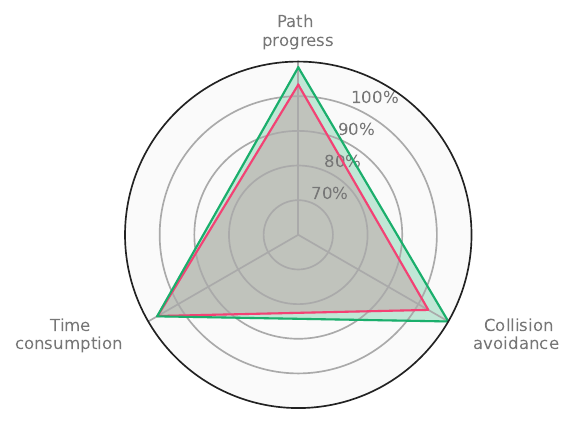}
\caption{Case 2}
\label{fig:case2_test}
  \end{subfigure}
  \begin{subfigure}[b]{0.32\linewidth}
\includegraphics[width=\textwidth]{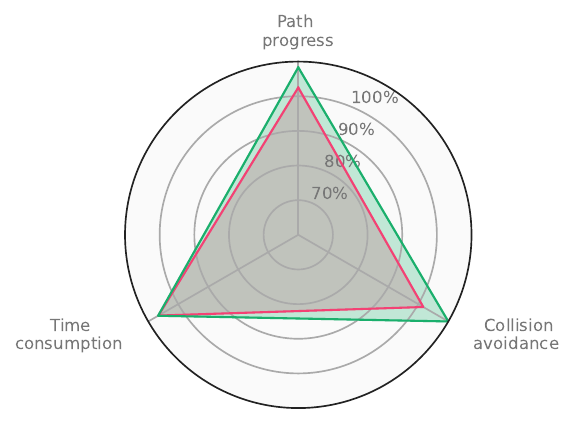}
\caption{Case 3}
\label{fig:case3_test}
  \end{subfigure}\\
  \begin{subfigure}[b]{\linewidth}
  \centering
\includegraphics[width=\linewidth]{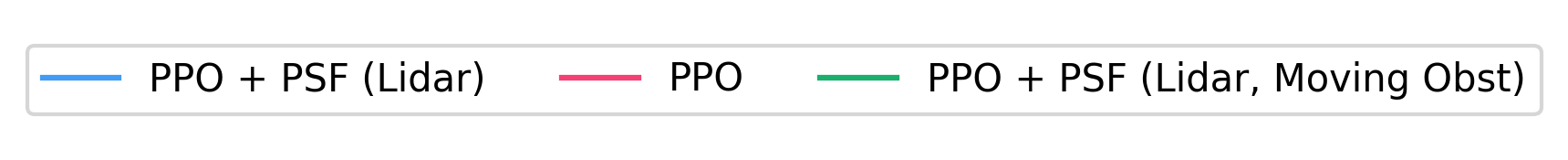}
  \end{subfigure}
\caption{Test results for Case 1, 2 and 3. In Case 2 and 3 the PSF was modified to have access to information about the moving obstacles in addition to the LiDAR obstacle detection. The PPO+PSF agent has a perfect score on path progress and collision avoidance in all three cases.}
\label{fig:testresults}
\end{figure}

After each agent was trained for 1 million timesteps, they were run for 100 additional episodes to evaluate the test performance. The results are visualized in the radar charts in Fig.~\ref{fig:testresults}. The PPO+PSF agents maintained perfect collision avoidance for all 3 cases, successfully completing every episode. The standard PPO agent did not achieve the same level of performance, scoring 98\%, 96\% and 95\% respectively in cases 1, 2, and 3 in collision avoidance. As expected, the PPO agent struggled more in the scenarios with the moving obstacles. 

\subsubsection{Case 4: Real environment}
\begin{figure}[h]
  \centering
  \begin{subfigure}[b]{0.32\linewidth}
    \includegraphics[width=\linewidth]{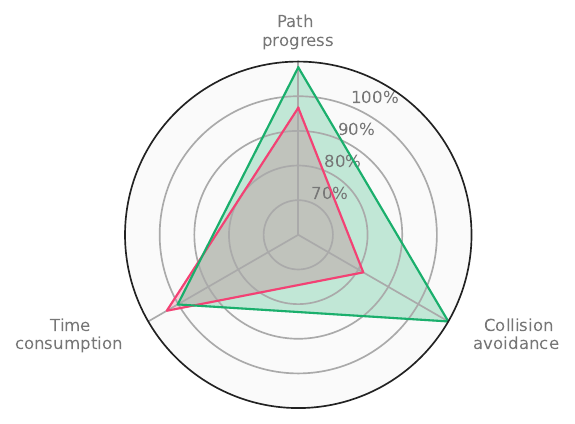}
    \caption{Trondheim}
    \label{subfig:Trondheim_radar_chart}
  \end{subfigure}
  \begin{subfigure}[b]{0.32\linewidth}
    \includegraphics[width=\linewidth]{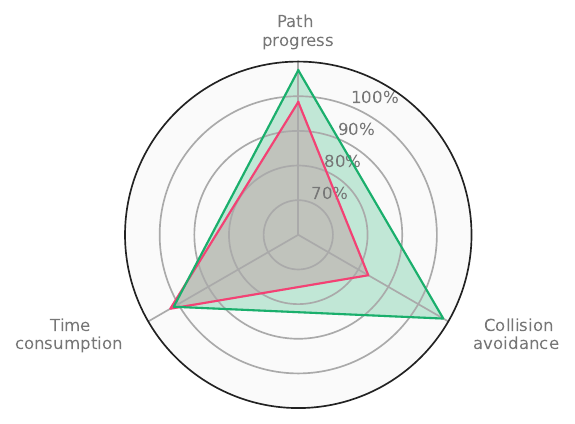}
    \caption{Agdenes}
    \label{subfig:Agdenes_radar_chart}
  \end{subfigure}
  \begin{subfigure}[b]{0.32\linewidth}
    \includegraphics[width=\linewidth]{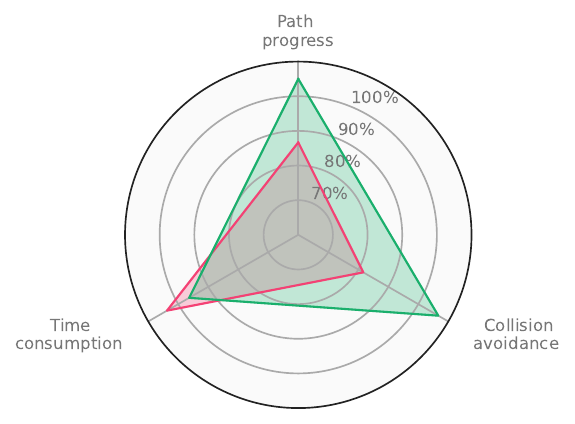}
    \caption{Sørbuøya}
    \label{subfig:Sorbuoya_radar_char}
  \end{subfigure}\\
    \begin{subfigure}[b]{\linewidth}
    \centering
    \includegraphics[width=0.3\linewidth]{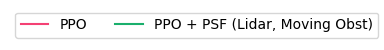}
  \end{subfigure}
\caption{Test results for the Trondheim, Agdenes, and Sørbuøya scenario. The PPO+PSF agent performs better on path progress and collision avoidance in all three cases, but it has a slightly higher time consumption.}
\label{fig:testradarTAS}
\end{figure}

Case 4 is the most challenging test case, with 3 real-world environments (Trondheim, Agdenes, Sørbuøya) of increasing difficulty. Prior to testing in these environments, the agents are trained for 2.000.000 timesteps on randomized scenarios generated according to case 3, with the PPO + PSF agent using both LiDAR-based collision avoidance and explicit moving obstacle collision avoidance. As can be inferred from Fig.~\ref{fig:testradarTAS}, in general, the agents perform worse in these environments. However, the discrepancy between the agents across the various performance metrics is also much more noticeable. For the Trondheim environment (Fig.~\ref{subfig:Trondheim_radar_chart}), PPO + PSF has a 0\% collision rate compared to 17\% for the standard PPO agent, at the expense of a significantly higher average cross-track error. The high average cross-track error for PPO + PSF was also inflated by a single episode in which the agent deviated significantly from the path, likely as a result of unsuccessfully trying to overcut a target ship traveling across it, resulting in the agent being stuck moving parallel to the target ship and deviating several hundred meters from the path. The PPO + PSF agent also uses more time on average, but this is largely a result of the standard PPO agent colliding, therefore, shortening the average episode duration. Looking at the average progress of the episodes, we see that PPO + PSF indeed manages to successfully reach the goal more often.

In the Agdenes environment (Fig.~\ref{subfig:Agdenes_radar_chart}), the PPO + PSF agent again performed significantly better than standard PPO, however, a single collision was registered, meaning that the predictive safety filter was not able to fully guarantee the safety of the agent in this environment. Looking at the scenario where the agent collided, the underlying cause seems to be the error between the predicted linear trajectory and the actual trajectory of the target ships in the vicinity of the agent. In this particular situation, a target ship was crossing the path in front of the agent at a safe distance. However, instead of following a predicted linear path, the target ship turned starboard in the direction of the agent, in accordance with the predescribed trajectory of the target ship sampled from AIS data.  As the ships were already close to each other when this maneuver occurred, the agent was not able to avoid collision with the given vessel dynamics and available thrust However, it should be mentioned that even though the trajectories of the target ships are taken from real AIS data, they themselves do not try to avoid collision in any way.

The Sørbuøya environment (Fig.~\ref{subfig:Sorbuoya_radar_char}) is the most difficult to navigate, which is reflected in the results. Again, the PPO + PSF agent has much lower collision rate, but still there were 2 registered collisions across the 100 episodes. Additionally, the standard PPO agent has a significantly shorter average episode duration, but again this is most likely a result of episodes being terminated early as a result of collisions. Another interesting result is that even though the PPO + PSF agent only collides twice, there were additional instances where it still did not reach the goal despite not colliding. Because the Sørbuøya environment consists of many densely packed islands, a situation can arise where the agent is stuck in a local minima. If the agent is surrounded by many small islands in close vicinity, it might not be possible to move closer to the goal without violating the minimum obstacle safety distance. If the paths to the left and right  are also blocked, the only viable strategy is to turn around and backtrack, moving away from the goal. This situation is much less common in the training environments, so the agent is unlikely to identify the solution.

\subsection{Ablation studies}
\subsubsection{Impact of PSF deviation penalty on agent safety awareness}
Since the predictive safety filter prevents collisions, the RL agent cannot rely on the collision penalty to learn safe behavior. Therefore, the collision risk penalty (section \ref{sec:collision_avoidance_reward}) and the PSF violation penalty (section \ref{sec:PSF_penalty}) provide the only signals for learning to avoid obstacles. To assess whether the agent still learns to behave safely while being aided by the PSF, and to compare the contribution of the different penalty terms, Agents with different penalty configurations where trained with the PSF activated for 1M timesteps, and then tested with the PSF deactivated across 100 randomized environments with static obstacles. The results are shown in table \ref{tab:ablation_PSF_penalty}.
\begin{table*}[h]
\centering
\caption{Collision avoidance rate with different configurations of penalty terms in the RL cost function. PSF used during training, but deactivated during testing}
\label{tab:ablation_PSF_penalty}
\begin{tabular}{|c| c c c c|}
\hline
 & \textbf{PSF penalty + Col. risk} & \textbf{Col. risk} & \textbf{PSF penalty} & \textbf{None} \\
\hline
\textbf{COLAV} & $99\%$ & $99\%$ & $94\%$ & $23\%$ \\
\hline
\end{tabular}
\end{table*}
Table \ref{tab:ablation_PSF_penalty} seems to indicate that the collision risk penalty (based on the LiDAR measurements) is sufficient for the agent to learn to avoid obstacles. The agent performs better when trained with only the collision risk penalty, compared with only being trained with the PSF penalty. This can be explained by the fact that the collision risk directly translates LiDAR measurements to a reward signal that promotes collision avoidance. When only the PSF penalty is used, the agent learns to associate the activation of the PSF (which occurs when a potential collision is imminent) with the corresponding LiDAR measurements, which indirectly promotes collision avoidance. However, Using only the PSF activation penalty is still drastically better than having no penalty terms at all, increasing the percentage of collision-free episodes from 23\% to 94\%.

\section{Conclusion and Future Work}
\label{sec:conclusionandfuturework}
We have provided a solution for autonomous control of an ASV combining a PSF with an RL agent. The method allows an RL agent to propose actions, with the PSF performing corrections in order to guarantee constraint satisfaction, which is vital for safety-critical systems. We demonstrated how the inclusion of a PSF significantly reduces the number of collisions during the testing and training of a state-of-the-art RL algorithm in various complex scenarios, some including environmental disturbances. The main conclusions can be itemized as follows:
\begin{itemize}
    \item Based on the simulation results, we demonstrated that the predictive safety filter is a promising strategy for ensuring safety in learning-based ship navigation and control. The PSF successfully managed to prevent the agent from colliding during all training and test episodes for each of the randomized test cases. Furthermore, the enhanced agent with PSF performed just as well as the baseline, requiring fewer training episodes to converge to a satisfactory performance level. Reengineering the PPO reward function to accommodate the predictive safety filter was relatively straightforward, requiring only that the collision penalty be replaced by a comparable PSF-activation penalty, and performing simple tuning. This indicates that the introduction of the PSF is not prohibitive with regard to the additional time spent re-designing the RL algorithm itself. Using state-of-the-art nonlinear model predictive control software, an average OCP solver runtime of less than 10 milliseconds was achieved, which is comfortably within the requirements of real-time application. Furthermore, the fast run-time of the optimization solver meant that the PSF-enhanced agent needed only slightly more time to complete the same number of simulation timesteps compared with the standard PPO agent.
    \item As expected, the predictive safety filter had the most impact in the initial training phase. During the first 100,000 timesteps of training, the standard PPO agent often registered a collision rate above 50\%, so using the PSF allows the learning agent to stay alive and collect much more experience from the initial episodes. This is advantageous not only from a safety perspective but also in terms of learning efficiency. In miniature-scale physical experiments, the cost of collision is not necessarily high. However, keeping the agent from colliding as long as possible can save a significant amount of time and labor that would otherwise be necessary to reset the episode and environment after a collision. For fully trained agents (trained for 1 to 2 million timesteps), the PPO + PSF agents behaved similarly to the standard PPO agents in the randomly generated environments, with close to the same average cross-track error and choosing similar paths most of the time. 
    \item All agents performed worse in real-world environments, although the PSF-enhanced agent still performed significantly better than the standard PPO agent. The collision avoidance rate and, consequently, the successful episode rate were significantly higher for PPO + PSF compared to standard PPO. Still, in a total of 300 simulations in real-world scenarios, there were 3 instances where the predictive safety filter was not able to prevent the collision. Identifying the exact reasons for the collisions is difficult due to the level of complexity in these environments. However, the fact that collisions happened even with the PSF enabled suggests that transferring from generated to real environments still poses numerous challenges, and that sufficient variation in the training phase is imperative to minimizing risk when applying the agents to real environments. The Sørbuøya environment also showed that, in some situations, the additional safety margins imposed by the predictive safety filter can hinder the forward progress of the agent, even though a feasible path exists. This indicates that the trade-off between safety and freedom of exploration must be considered carefully when designing the PSF. Despite this, the overall results suggest that predictive safety filters can significantly improve safety when RL-based autonomous vessels are deployed in real environments and thereby increase the viability of reinforcement learning in marine navigation and control.
\end{itemize}

Further development is needed to fully bridge the gap from simulation to performing real experiments. In a real setting, factors such as measurement noise, modeling error, and sensor failure must be handled rigorously to fully ensure safety and efficiency. To address this, a key area for future research is to extend the predictive safety filter formulation to a \textit{robust predictive safety filter} \cite{Wabersich2021aps} The variation in randomly generated environments, in terms of the distribution and complexity of static and dynamic obstacles, can be increased to better prepare the RL agents for the wide range of scenarios they encounter in the real world. Another interesting topic that was not addressed in this work is the interaction and possible cooperation between multiple autonomous vessels. In a realistic setting, all involved ships cooperate to minimize collision risk and maximize efficiency. By deploying multiple agents simultaneously in the same environment, collective behaviors can be studied, and multi-agent strategies for collision avoidance can be developed.

\section*{Acknowledgments}
This work is part of SFI AutoShip, an 8-year research-based innovation center. 
In addition, this research project is integrated into the PERSEUS doctoral program. 
We want to thank our partners, including the Research Council of Norway, under project number 309230, and the European Union’s Horizon 2020 research and innovation program under the Marie Skłodowska-Curie grant agreement number 101034240. 
\bibliographystyle{AR}
\bibliography{references}
  
\appendix

\end{document}